\RequirePackage{fix-cm}
\documentclass[twocolumn]{svjour3}          
\smartqed  
\usepackage{graphicx}
\usepackage[colorlinks=true,urlcolor=blue,citecolor=blue,linkcolor=blue,bookmarks=true]{hyperref}
\usepackage{xurl}
\usepackage{booktabs}       
\usepackage{amsfonts}       
\usepackage{nicefrac}       
\usepackage{microtype}      
\usepackage{flushend}
\usepackage{multirow}
\usepackage{multicol}
\usepackage{amssymb}
\usepackage{amsmath}
\usepackage{subcaption}
\usepackage{colortbl}
\usepackage{changepage}
\usepackage{soul, color}
\usepackage[utf8]{inputenc}
\usepackage[dvipsnames]{xcolor}
\usepackage{xspace}
\usepackage{threeparttable}
\usepackage{bm}
\usepackage{wasysym}
\usepackage{pifont}
\usepackage{fancyvrb}
\usepackage{amsbsy}

\usepackage{bbding}
\usepackage{natbib}

\definecolor{Gray}{gray}{0.9}
\definecolor{beaublue}{RGB}{223,235,241}
\definecolor{audiogreen}{RGB}{234,241,230}
\definecolor{visualblue}{RGB}{231,240,244}
\definecolor{fusionorange}{RGB}{253,236,215}
\definecolor{mydarkblue}{rgb}{0,0.08,0.45}

\makeatletter
\DeclareRobustCommand\onedot{\futurelet\@let@token\@onedot}
\def\@onedot{\ifx\@let@token.\else.\null\fi\xspace}
\def\eg{\emph{e.g}\onedot} 
\def\ie{\emph{i.e}\onedot} 
 
\def\etc{\emph{etc}\onedot}

\makeatother

\journalname{International Journal of Computer Vision}
\begin{document}

\title{Day2Dark: Pseudo-Supervised Activity Recognition beyond Silent Daylight}


\author{Yunhua Zhang \and Hazel Doughty \and Cees G.M. Snoek
}


\institute{
            University of Amsterdam, Netherlands \\
            \email{\{y.zhang9,hazel.doughty,cgmsnoek\}@uva.nl} \\
}

\date{Received: date / Accepted: date}

\maketitle

\begin{abstract}
This paper strives to recognize activities in the dark, as well as in the day. We first establish that state-of-the-art activity recognizers are effective during the day, but not trustworthy in the dark. The main causes are the limited availability of labeled dark videos to learn from, as well as the distribution shift towards the lower color contrast at test-time. To compensate for the lack of labeled dark videos, we introduce a pseudo-supervised learning scheme, which utilizes easy to obtain unlabeled and task-irrelevant dark videos to improve an activity recognizer in low light. As the lower color contrast results in visual information loss, we further propose to incorporate the complementary activity information within audio, which is invariant to illumination. 
Since the usefulness of audio and visual features differs depending on the amount of illumination, we introduce our `darkness-adaptive' audio-visual recognizer. 
Experiments on EPIC-Kitchens, Kinetics-Sound, and Charades demonstrate our proposals are superior to image enhancement, domain adaptation and alternative audio-visual fusion methods, and can even improve robustness to local darkness caused by occlusions. Project page: \url{https://xiaobai1217.github.io/Day2Dark/}
\end{abstract}
 
\keywords{Video analysis \and Audio-visual learning}

\section{Introduction}

The goal of this paper is to recognize activities like `someone is opening window' or `running' captured in videos under challenging low-illumination conditions, while maintaining recognition in daylight. Activity recognition has been studied for decades with tremendous progress \eg, \cite{marszalek09,kuehne2011hmdb,slowfast,lee2021crossattentional}, most recently by learning activity classes from labeled video examples \cite{girdhar2022omnivore,wu2022memvit,yan2022multiview}. 
Despite these advancements, we observe the large majority of activity recognizers are trained on well-lit videos, as this is what current datasets contain. 
For example, only 4.4\% and 1.9\% of the videos in the widely used Kinetics-400~\citep{Kinetics} and EPIC-Kitchens~\citep{EPIC} datasets are recorded in the dark.  Due to the few dark training videos available and the loss of information from the low color contrast, state-of-the-art models struggle to generalize to low-light conditions at test-time. This inability hurts a wide range of night-time applications, \eg, smart homes, self-driving cars, security cameras and wildlife monitoring. Thus, in this work, we propose to enhance an activity recognition model with the ability to generalize to nighttime video conditions.

Several computer vision solutions for activity recognition in the dark already exist. 
For instance, \cite{action02mcf} introduce a correlation filter for dark scenarios and \cite{xu2020arid} collect a dataset with videos recorded in the dark. They deal with low light conditions by training on labeled dark videos. \cite{gao2016infar} construct a dataset captured by infrared sensors, which are invariant to the illumination. Correspondingly, \cite{liu2018global} and \cite{jiang2017learning} propose representation learning techniques for infrared data. However, the model learning of all these methods requires manually labeled dark videos or infrared data, which are difficult to obtain and scale, and may work in low light only, \eg~\cite{action02mcf,xu2020arid}. Inspired by prior works using pseudo-supervision~\citep{lee2013pseudo, gavrilyuk2021motionaugmented, doughty2022you}, we equip activity recognizers with dark recognition ability by learning from widely available unlabeled dark videos.

An alternative approach to recognition in the dark is image enhancement, \eg,~\cite{zhang2019kindling,ma2022toward,zhou2022lednet,jin2022unsupervised}. However, such methods introduce color distortions and discontinuities between video frames, which are harmful for activity recognition. The additional computation time needed by these methods also cannot be ignored. We instead make use of the illumination-invariant properties of sound, which contains rich activity information. While doing so, it is of importance to note that standard audio-visual fusion methods, such as~\cite{nagrani2021attention,lee2021crossattentional,gabeur2020multi}, are limited by their (unintended) focus on daylight. This daylight-focus makes audio-visual correspondences harder to establish under low-light conditions. We thus propose a darkness-adaptive audio-visual recognizer which learns to alter the fusion based on a video's illumination.

As our first contribution, we establish the \textit{day2dark gap} (Section~\ref{sec:day2darkgap}), which refers to the difference in activity recognition robustness between day and dark videos. We find that popular current video datasets for activity recognition are daylight-focused, and existing models, both convolutional and transformer-based, trained on these datasets struggle to recognize activities in the dark. 
Our analysis shows that the model degradation results from the lack of dark labeled training videos, the visual distribution shift under low illumination and the information loss caused by the lower color contrast. 

We propose a method to reduce the day2dark gap which is summarized in Section~\ref{sec:overview}. Our method consists of two key components, one focusing on the model supervision and one on the model architecture. 

\begin{figure}[t!]
\centering 
\includegraphics[width=0.95\linewidth,height=0.9\linewidth]{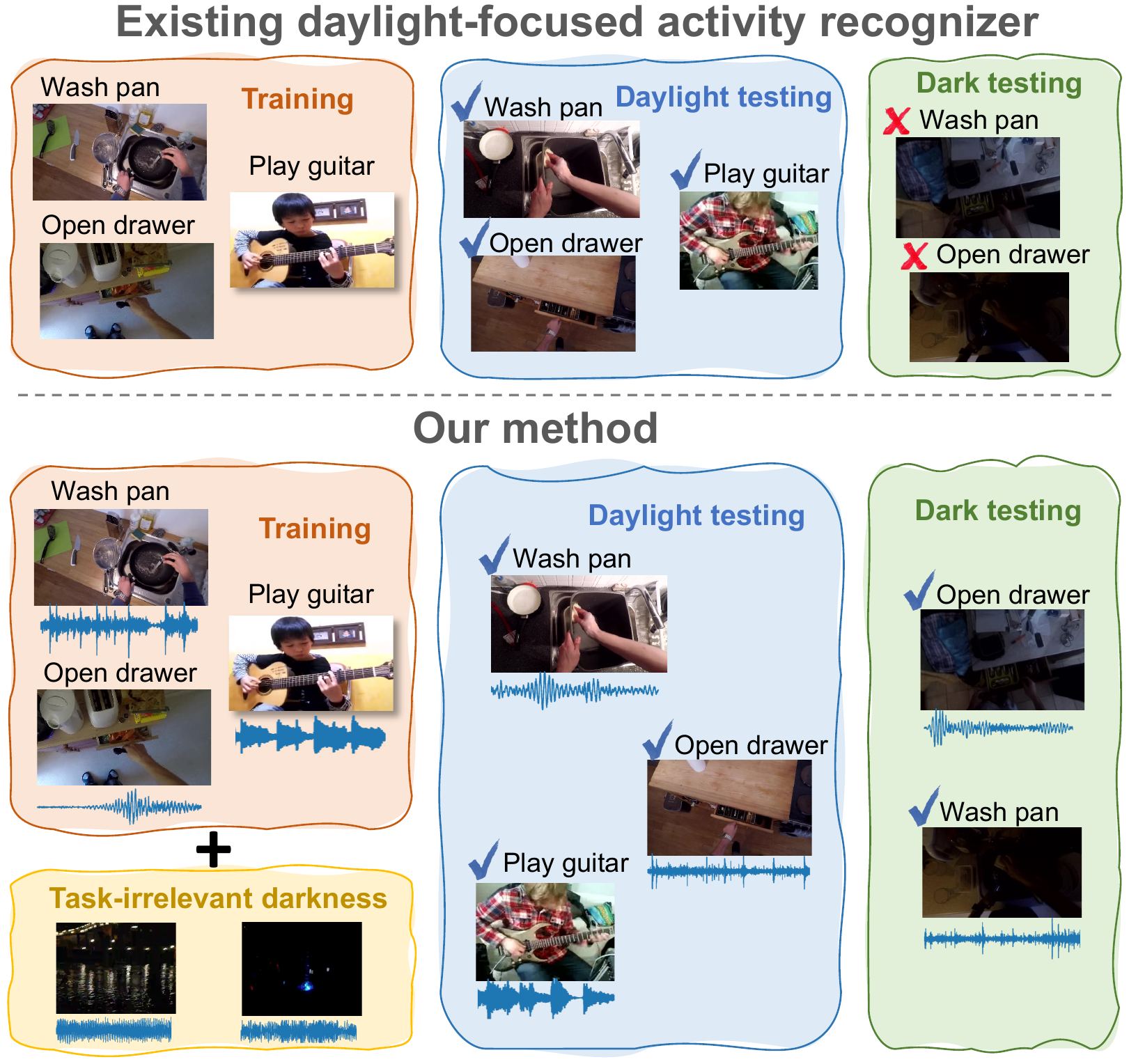}
\vspace{-0.5em}
\caption{
Activity recognizers (unintentionally) focus on well-lit videos during training. By doing so, they fail during deployment in the dark due to the distribution shift and low contrast at test-time. Our proposed method adapts to darkness without requiring task-relevant dark videos to learn from and adaptively fuses appearance with the illumination-invariant sound information according to their usefulness in the current light condition.
}
\label{fig:1st_figure}
\end{figure}

To compensate for the lack of labeled dark data, we propose a pseudo-supervised learning strategy as our second contribution (Section~\ref{sec:supervision}) that utilizes \textit{unlabeled} dark videos, which do not need to contain target activities (see Fig.~\ref{fig:1st_figure}). 
Combined with labeled daylight videos in existing datasets, the trained activity recognizer, either visual-only or audio-visual, is able to generalize beyond daylight videos to the dark data distribution. 

To compensate for the information loss by the low color contrast, we include illuminance-invariant information from sound. 
However, we find that directly fusing audio and visual modalities as existing methods~\citep{lee2021crossattentional,tian2020unified,shvetsova2022everything} leads to limited performance improvement in the dark, since the visual distribution shift makes the audio-visual correspondences harder to establish. Thus, our third contribution (Section~\ref{sec:adaptive_recognizer}) is introducing a darkness-adaptive audio-visual recognizer that analyzes the clarity of a video's visual features and accordingly reduces the distribution shift to find better audio-visual correspondences under different illuminations.

Lastly, we demonstrate the advantages of our learning strategy, our darkness-adaptive recognizer and the value of sound in low light video environments (Section~\ref{sec:experiments}). 
Results show that our proposed method can be used with different backbones (either convolutional- or transformer-based) and equips an activity recognizer with the ability to generalize to low-lit conditions even without any labeled dark videos in training. We further find our proposed model is more effective for activity recognition in the dark than recent image enhancement, domain adaptation and audio-visual fusion methods. Finally, we also find our model increases robustness to occlusions during daylight. Before detailing the day2dark gap and our methodology, we first embed our work more thoroughly in related work.

\section{Related Work}
\label{sec:related_work}

We begin our review by presenting previous works dealing with activity recognition under distribution shift in Section~\ref{sec:related_distribution}. Next, we briefly review methods for computer vision tasks in dark environments in Section~\ref{sec:related_dark}, including both image- and video-focused approaches. 
Then, in Section~\ref{sec:related_enhancement}, we introduce prior image enhancement methods that can improve color contrast, recover texture and remove noise for low-light image and have the potential to help low-light vision tasks. 
Finally, we summarize and discuss existing audio-visual fusion methods for daylight videos in Section~\ref{sec:related_sound}.

\subsection{Activity Recognition Under Distribution Shift}
\label{sec:related_distribution}

Some approaches for activity recognition adapt to visual distribution shifts caused by scenery~\citep{EPIC,munro2020multi}, viewpoint~\citep{sigurdsson2018actor} or actor~\citep{zhang2022audio}. 
Several works adopt adversarial learning to align features between the source and the target domains. 
For instance, \cite{jamal2018deep} and \cite{munro2020multi} directly penalize domain-specific features with an adversarial loss at every time stamp. 
\cite{chen2019temporal} explicitly attend to the temporal dynamics using domain discrepancy for more effective domain alignment by adversarial learning. 
\cite{choi2020shuffle} combine a domain adversarial loss with an auxiliary task to encourage learning of representations which focus on the humans and objects involved in the actions. 
\cite{pan2020adversarial} match the distributions of temporally aligned action features between source and target domains using a cross-domain co-attention mechanism and the proposed model is optimized in the adversarial learning manner. 
Self-supervised learning objectives are also widely used to project features of different domains into a common feature space. 
For example, \cite{munro2020multi} predict whether the input RGB and optical flow modalities come from the same video. \cite{choi2020shuffle} predict the temporal order of video clips to focus more on the action and less on the scene context. Both \cite{song2021spatio} and \cite{iccv2021videoadaptation} adopt contrastive learning. While \cite{song2021spatio} propose spatio-temporal contrastive learning to align both the clip-level and the video-level representation between the source and the target domains, 
\cite{iccv2021videoadaptation} apply cross-modal contrastive learning on RGB and optical flow modalities. 
Both \cite{yang2022interact} and \cite{zhang2022audio} combat domain shift by complementary information from sound. 
While \cite{yang2022interact} exchange complementary information across modalities, \cite{zhang2022audio} utilize the domain-invariant activity sounds to reduce the distribution shift in visual features. 
A few works~\citep{xu2021partial,xu2022learning,xu2021multi} also investigate distribution shift from low illumination and adapt from day to dark. 
However, domain adaptation methods need unlabeled target domain videos that contain activities of interest during model learning, while we only require videos to be low-light, a considerable practical advantage.

\subsection{Recognition in the Dark} 
\label{sec:related_dark}

Recently, there has been a surge in research interest with regards to computer vision tasks in dark environments. 
Several works focus exclusively on images. 
For instance, \cite{crowdcounting} improve crowd counting accuracy under low illumination and occlusion conditions by modulating the visual features by audio. 
\cite{neumann2018nightowls} demonstrate state-of-the-art pedestrian detectors do not perform well at night, even when specifically trained on night data, and there is a clear gap in accuracy between day and night detections. 
For semantic segmentation, \cite{sun2019see} use generative models to convert nighttime images into the daytime and also do the inverse to obtain more nighttime training data. 
While \cite{sakaridis2019guided} collect paired nighttime and daytime images to facilitate the research, \cite{gao2022cross} use these data to develop a domain adaptation method. 

There are also some works for vision tasks on videos recorded in the dark. 
While \cite{Chen_2019_ICCV,xu2020arid,ye2022unsupervised,jiang2019learning} introduce dark video datasets for visual model learning, \cite{gebhardt2018camel} collect infrared video data which are invariant to the illumination. 
An early work by \cite{action02mcf} proposes a space-time correlation filter for dark scenarios. 
The recent work by \cite{luo2023similarity} maximizes the feature similarity between the darkened images and their normal-light counterparts for better model adaptation. 
Furthermore, several works demonstrate the benefit of sound in the dark. 
\cite{gan2019self} and \cite{valverde2021there} utilize sound for object tracking, while \cite{xdviolence} and \cite{zhang2021repetitive} obtain considerable performance improvement in the dark for violence detection and repetition counting by the illumination-invariant audio information. Inspired by these works, we go one step further and propose to adaptively fuse sight and sound depending on their usefulness under low light. In contrast to existing solutions for activity recognition, which require large amounts of labeled dark videos, ours learns from labeled daylight videos and unlabeled dark videos which are both widely available.

\subsection{Image Enhancement} 
\label{sec:related_enhancement}

Recognition in the dark can also be achieved by first applying image enhancement. Gamma Intensity Correction~\citep{poynton2012digital} simply brightens an image through pixel intensity, while recent works learn deep networks to enhance the color contrast, recover texture and remove noise for low-light images. For instance, \cite{wang2018gladnet} design a global illumination-aware and detail-preserving network. 
\cite{wei2018deep} combine the Retinex theory with CNNs to estimate the illumination map and enhance the low-light images. 
\cite{zhang2019kindling} design a similar network but connect the feature-level illumination and reflectance in the decomposition step.
They later present an improved version~\citep{zhang2021beyond} with a mutli-scale illumination attention module to alleviate visual defects, \eg non-uniform spots and over-smoothing. 
\cite{liu2021retinex} build a Retinex-inspired unrolling framework with architecture search, while \cite{wu2022uretinex} unfold an optimization problem into a learnable network to decompose a low-light image into reflectance and illumination layers. 
\cite{ma2022toward} establish a cascaded illumination learning process with weight sharing to handle the low-light enhancement task. 
\cite{jin2022unsupervised} introduce an unsupervised method that integrates a layer decomposition network and a light-effects suppression network. 
As \cite{zhou2022lednet} utilize data synthesis for model learning, \cite{wang2022enhancement} exploit local information to benefit both under- and over-exposure regions. While effective for images, applying enhancement to videos introduces color distortions and discontinuities between frames, which are harmful for activity recognition. Besides, these methods also add a non-negligible computation cost. We instead recognize activities with the aid of illumination-invariant sound. 


\subsection{Sound for Activity Recognition.} 
\label{sec:related_sound}

Several prior works already demonstrate the benefit of sound for activity recognition. 
For instance, \cite{korbar2018cooperative} learn general and effective models for both audio and video analysis from self-supervised temporal synchronization. 
Both \cite{gao2020listen} and \cite{korbar2019scsampler} reduce computation of visual features by previewing audio. 
Cross-modal attention is also effective in activity recognition~\citep{nagrani2021attention}, event-localization~\citep{lee2021crossattentional,tian2018audio,wu2019dual}, video captioning~\citep{rahman2019watch,tian2018attempt,wang2018watch}, video retrieval~\citep{gabeur2020multi,shvetsova2022everything} and video parsing~\citep{tian2020unified,wu2021exploring}. 
For instance, \cite{gabeur2020multi} present a multi-modal transformer to jointly encode the different modalities in video, which allows each of them to attend to the others. 
\cite{tian2020unified} propose hybrid attention with multimodal multiple instance learning for weakly-supervised audio-visual video parsing. 
\cite{nagrani2021attention} introduce a multi-modal bottleneck transformer for effective audio-visual fusion trained in the supervised setting. 
Inspired by this, \cite{lin2023vision} adapt a pre-trained vision transformer for audio-visual fusion by bottleneck representation. 
\cite{lee2021crossattentional} fuse audio and visual features by multi-stage cross-attentional audio-visual fusion for localizing and classifying actions in videos. 
\cite{li2023decoupled} propose a cross-modal knowledge distillation approach to enhance the discriminative features of each modality. 
All these works learn their audio-visual fusion with datasets primarily containing daylight videos. 
Different from these works, our darkness-adaptive fusion adjusts to the illumination allowing our model to be better suited for day and dark video conditions. 

\section{The Day2Dark Gap}
\label{sec:day2darkgap}
In this section, we establish that video datasets for activity recognition are daylight-focused. Consequently, we demonstrate that models, trained on these datasets, are unable to generalize well to dark scenarios. We call this difference in robustness between day and dark videos the Day2Dark Gap. 

\subsection{Identifying Dark Videos}

To analyze the content of activity recognition datasets and investigate how performance changes with illumination, we first need to compute the illuminance per video. Following~\cite{anderson1996proposal} we obtain an illuminance value $Y$ for each video frame with height and width $H{\times} W$:
\begin{equation}
\label{eq:y_compute}
Y= \frac{
     \sum_{j=1}^{H{\times}W} (0.299R_j+0.587G_j+0.144B_j)}
     {H{\times}W}.
\end{equation}
$R_j$, $G_j$, and $B_j$ indicate the intensities of the red, green and blue channels of the \textit{j}th pixel. 
We consider a video's illuminance to be the average across all frames and we observe that videos captured in the dark usually have $Y{\leq}40$.


\begin{figure*}[t!]
\centering 
\includegraphics[width=\linewidth]{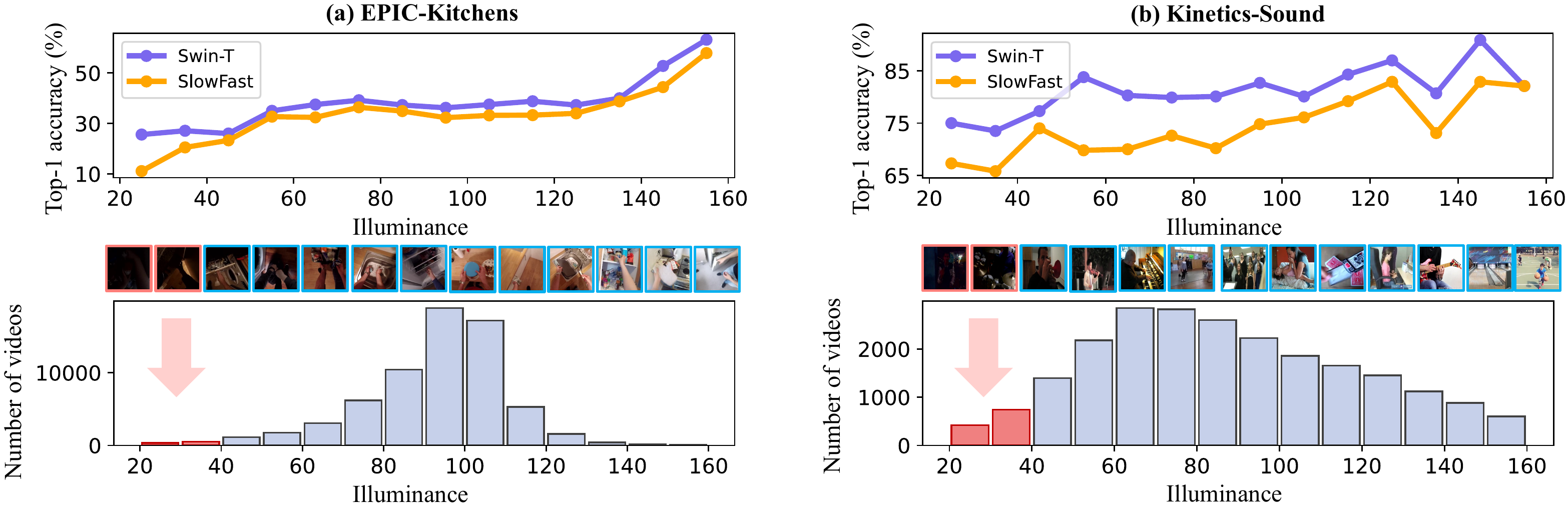}
\vspace{-1.5em}
\caption{\textbf{Performance Variation and Training Set Size Across Illuminations} on (a) EPIC-Kitchens and (b) Kinetics-Sound. Current transformer (Swin-T~\citep{liu2021swin}) and CNN (SlowFast~\citep{slowfast}) models experience severe performance drops in low-illumination. 
}
\label{fig:2nd_figure_epic}
\end{figure*}
%

\subsection{Datasets}
In current activity recognition datasets, only a small percentage of videos are captured in low-light. For instance, Kinetics-400 has only 4.4\% of videos with illuminance $Y{\leq}40$, EPIC-Kitchens has 1.9\%, Moments-in-Time 4.9\%, Charades 3.6\%, Kinetics-Sound 8.3\% and ActivityNet 3.2\%. Instead, the large majority of clips in these datasets are daylight videos where the activity can be seen clearly. 

As dark videos are the minority in training, we are in danger of not being able to perform activity recognition in the dark with current models, necessary for applications in smart homes and video surveillance. Since dark videos are also the minority in testing, we have no idea how models perform in the dark as test sets are evaluated as a whole. 
To shed an empirical light on the matter, we perform an analysis using two widely used activity recognition datasets: EPIC-Kitchens~\citep{EPIC} and Kinetics-Sound~\citep{KineticsSound}. It reveals how activity recognition models perform under different illumination conditions.

\subsection{Analysis}
\begin{figure*}[t!]
\begin{subfigure}{\linewidth}
\includegraphics[width=\linewidth]{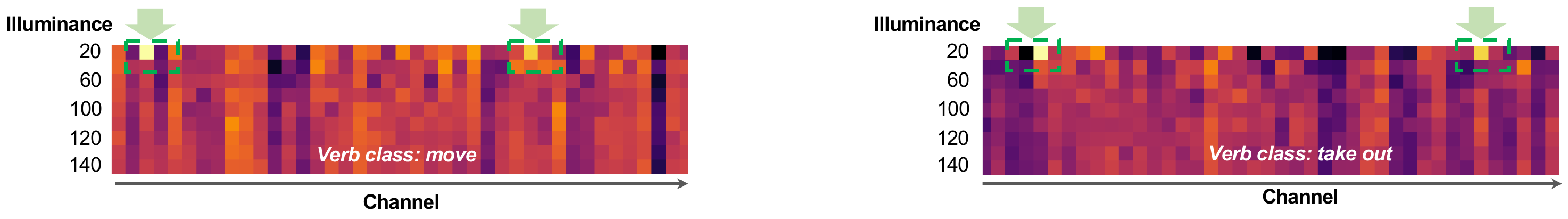}
\caption{\textbf{Swin-T}}
\end{subfigure}
\bigskip
\begin{subfigure}{\linewidth}
\includegraphics[width=\linewidth]{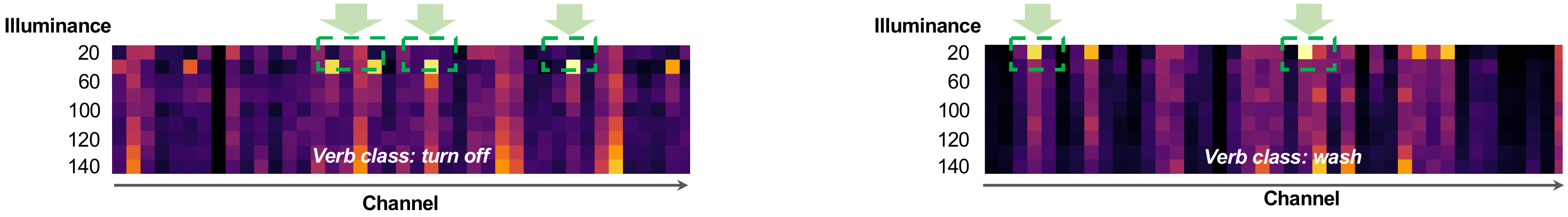}
\caption{\textbf{SlowFast}}
\end{subfigure}
\vspace{-3em}
\caption{\textbf{Visual Feature Distribution Shift in the Dark}. We show the average activation for 50 random feature channels over different illuminances for different activity classes. Visual features are similar across different illuminances, but change when the illuminance is lower than 40. }
\label{fig:visual_distribution}
\end{figure*}

To understand how well current models generalize to dark videos, we examine activity recognition performance across illuminations. 
Fig.~\ref{fig:2nd_figure_epic} shows these results for EPIC-Kitchens and Kinetics-Sound with Swin-T~\citep{liu2021swin} and SlowFast~\citep{slowfast} backbones. 
For both datasets and backbones, we observe severe performance drops with illuminance $Y{\leq}40$. 
One reason is the few dark videos seen in training, meaning models struggle to generalize. %
This explains why the performance drop is larger on EPIC-Kitchens, where the percentage of dark training videos is lower than in Kinetics-Sound. 
However, models can generalize well to high-illuminance videos, which also have few examples in training. 
To understand this phenomenon, we visualize the activation of visual feature channels under various illuminations in Fig.~\ref{fig:visual_distribution} for both Swin-T and Slowfast. 
Specifically, we randomly pick 50 channels for each class and then compute the average feature activation across all videos in each illumination interval. 
We observe that feature activations of both models are often similar for the different illuminances with $Y{>}40$, however, some feature channels change dramatically with $Y{\leq}40$. 
For example, the verb class \textit{take out} with Swin-T, shown in the top right corner of Fig.~\ref{fig:visual_distribution}, has two channels where there is significant activation change compared to daylight conditions as highlighted by the green boxes. 
In contrast, when the illuminance becomes high with $Y{>}120$, the feature activations are similar to normal light conditions with $40{\leq}Y{\leq}120$. 
Some features even become more pronounced in high light conditions with larger activations, for instance for the \textit{turn off} class with SlowFast in the lower left. This is likely the cause for accuracy improvement at higher illuminances. 
These phenomenons are common for different activity classes with both the transformer (Swin-T) and the convolutional (SlowFast) backbones. 
Hence, there is a visual distribution shift in the dark partly caused by the loss of information from the low color contrast, and this shift results in model degradation. 
We conclude that the failures of activity recognizers in the dark stem from the visual distribution shift of dark videos which the model struggles to generalize to without sufficient dark training videos.

\subsection{Conclusion}
While existing vision-focused activity recognizers achieve good performance on daylight videos, they suffer from degradation in the dark. This is a wide-spread issue which occurs for multiple types of backbone on multiple datasets. The lack of any performance-drop on the, also rare, high illuminance videos reveal the degradation in the dark is not only caused by the lack of training data, but also by the visual distribution shift from the lower color contrast. 

\section{Reducing the Day2Dark Gap}
\label{sec:overview}

\subsection{Task Definition}

Our goal is to remove the day2dark gap in activity recognition and accurately recognize activities in both day and dark conditions. 
We have a set of $n$ videos $\mathcal{S}{=}\{(x^i, y^i) | i \in [1,n] \}$, where $x^i$ is a video and $y^i$ is its label. 
These videos mostly have high illumination, but a small proportion may have low illumination as highlighted in Section~\ref{sec:day2darkgap}. 
Since there are unlabeled dark videos widely available, the task can be relaxed to access any such videos freely. 
Consequently, we introduce an additional set of $m$ dark videos $\mathcal{U}{=}\{x^j | j \in [1,m]\}$, which have an average illuminance below a threshold, \ie, $Y{\leq}t$. 
These videos are unlabeled and do not need to contain activities we aim to recognize, \ie, they are task-irrelevant. 
The goal is to train a model to correctly predict the activities in test set $\mathcal{T}$, which contains both day and dark videos. 
Note our task is distinct from domain adaptation~\citep{wang2018deep,yang2022interact,zhang2022audio} which requires the unlabeled set to contain the same activities as the labeled set. It is also distinct from domain generalization~\citep{wang2022generalizing} as we have unlabeled dark videos in training.

\subsection{Method Overview}
To effectively utilize the unlabeled dark videos, we introduce supervision beyond daylight to train an activity recognizer for both day and dark scenarios. 
Our proposed supervision is applicable to different activity recognizers, \eg existing visual-only or audio-visual models. 
As a result, the trained activity recognizer can generalize beyond daylight to low-lit conditions. 
To further address the information loss caused by the lower color contrast in the dark, we propose a new audio-visual model to effectively utilize the complementary information within sound under different illuminations. 
We find that existing audio-visual fusion are ill-equipped for low light conditions as the visual distribution shift makes audio-visual correspondences harder to find. 
Therefore, our proposed darkness-adaptive audio-visual recognizer reduces the visual distribution shift and adaptively fuses audio and visual features according to the illumination. We first detail our proposed supervision beyond daylight in Section~\ref{sec:supervision}, before describing our darkness-adaptive audio-visual recognizer in Section~\ref{sec:adaptive_recognizer}.

\section{Supervision beyond Daylight}
\label{sec:supervision}

\subsection{Overview}
To improve an activity recognizer in the dark, we propose supervision beyond daylight that effectively uses unlabeled dark videos with two-stage training. 
Our supervision can be used to train any visual-only or audio-visual activity recognizer. 
In stage 1, we train the activity recognizer with pseudo-supervision on the unlabeled dark videos and full supervision on labeled daylight videos. 
This allows the model to generalize beyond daylight videos to the dark data distribution. 
In stage 2, we finetune the model for recognition in the dark by mixing labeled day and unlabeled dark videos with our proposed day2dark-mix.

\subsection{Stage 1: Pseudo-Supervised Day2Dark Learning}

To train our model with the unlabeled dark videos $U$, we need pseudo-labels. 
We could use the predictions of an activity recognizer trained on the labeled videos $S$. %
However, the unlabeled dark videos do not contain activities of interest, meaning the predictions will be uninformative and may be similar for many videos meaning it will not provide a useful training signal. 
Thus, we obtain pseudo-labels for the unlabeled dark videos with models trained on self-supervised auxiliary tasks, such as video-text matching~\citep{miech2019howto100m,bain2021frozen}. 
While the predictions for this task are different from our goal (activity recognition), they provide a training signal to learn distinguishing features in the dark. 
To prevent over-specialization, we use multiple auxiliary tasks simultaneously. 

\begin{figure}[t!]
\centering 
\includegraphics[width=\linewidth]{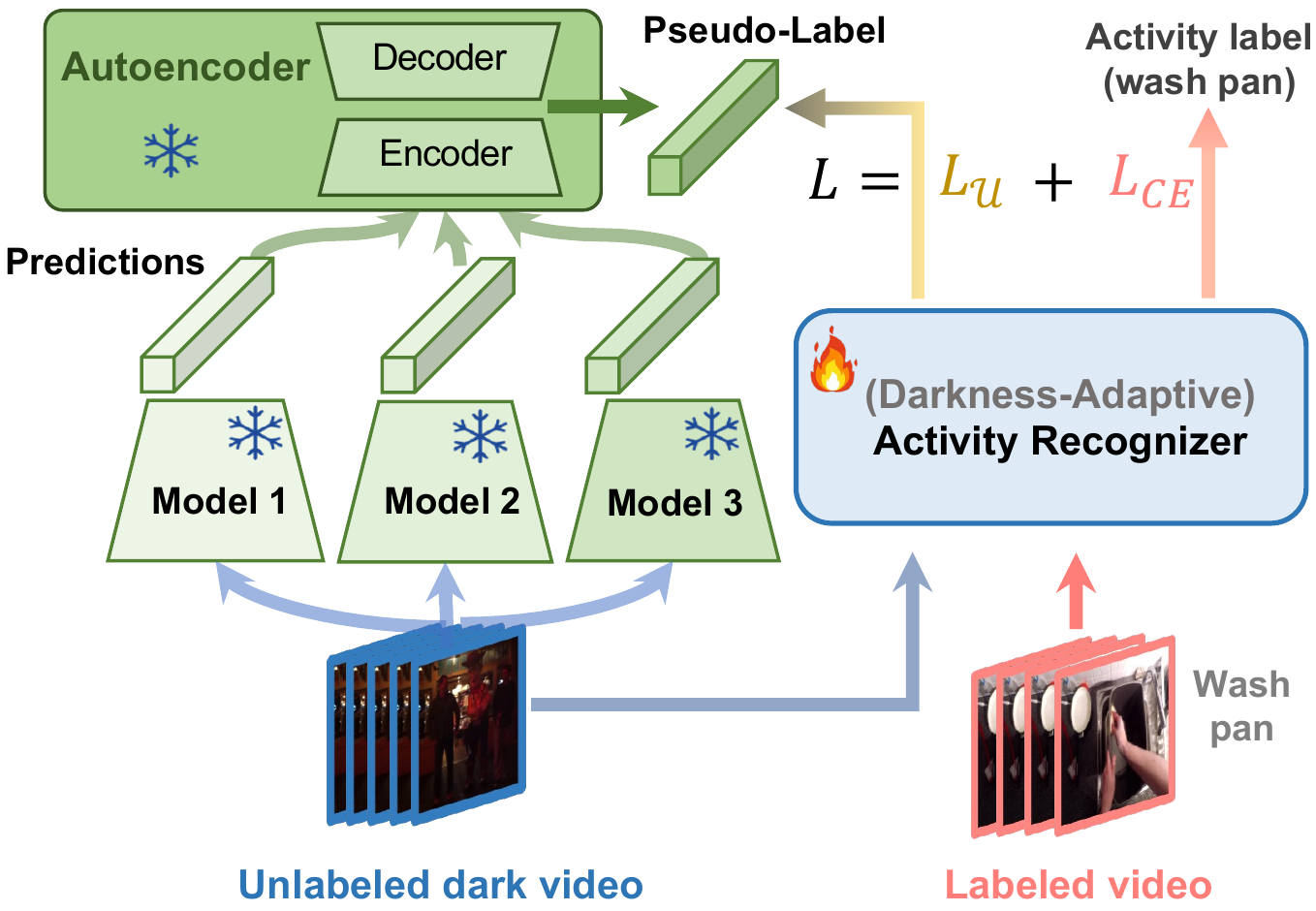}
\vspace{-1.5em}
\caption{
\textbf{Stage 1: Pseudo-Supervised Day2Dark Learning.} 
For each unlabeled dark video, we obtain predictions from auxiliary tasks. 
To prevent overfitting to these tasks, the predictions are encoded by an autoencoder. 
The activity recognizer is trained to produce the latent representation for unlabeled videos and predict the activities in labeled videos. 
}
\label{fig:supervision}
\end{figure}

Our pseudo-supervised day2dark learning is shown in Fig.~\ref{fig:supervision}. 
Specifically, we collect $m$ off-the-shelf self-supervised models $\{\phi_1(\cdot), \cdots, \phi_m(\cdot)\}$ for $m$ auxiliary tasks, such as video-text retrieval and sound source localization. 
Note that we use the same modalities for the auxiliary tasks as the activity recognizer, \ie, audio and visual or visual-only. 
While these self-supervised models may have extra modalities, \eg text, they are not used. 
For each unlabeled dark video, we obtain predictions from the $m$ models, \ie, $\{\mathbf{p}_1, \cdots, \mathbf{p}_m | \mathbf{p}_i \in \mathbb{R}^{d_m}\}$, where $d_m$ is the task's output dimension, \eg, the embedded feature dimension. 
We could use these predictions directly as pseudo-labels, however, they can have many dimensions, \eg activation maps, 
which causes overfitting to the auxiliary tasks. 
We instead use a compact, more abstract representation of the pseudo labels. 
Specifically, we give the concatenated predictions $[\mathbf{p}_1, \cdots, \mathbf{p}_m]$ to an autoencoder $\Phi(\cdot)$, which is trained in advance with an $L_1$ loss to reconstruct the input from a latent bottleneck representation. 
This latent provides us with a compact pseudo-label representation, $\mathbf{q}^j\in \mathbb{R}^{64}$. This becomes the target label for unlabeled video $x^j$ and is used in the loss:  
\begin{equation}
    L_{\mathcal{U}} = \sum_{j=1}^U \mathrm{dist}(\hat{\mathbf{q}}^j, \mathbf{q}^j),
\label{eq:L_U}
\end{equation}
where $\mathrm{dist}$ is a distance function, we use L1, and $\hat{\mathbf{q}}^j$ is an output of our activity recognizer optimized to be close to $\mathbf{q}^j$. 
For labeled videos $S$, we use a cross-entropy loss, $L_{CE}$, to also learn features relevant to activity recognition. The overall loss is:
\begin{equation}
    L = L_{CE} + \lambda L_{\mathcal{U}},
\label{eq:pseudo_supervised}
\end{equation}
where $\lambda$ is a weighting factor to balance the two losses. 

\begin{figure}[t!]
\centering 
\includegraphics[width=0.9\linewidth]{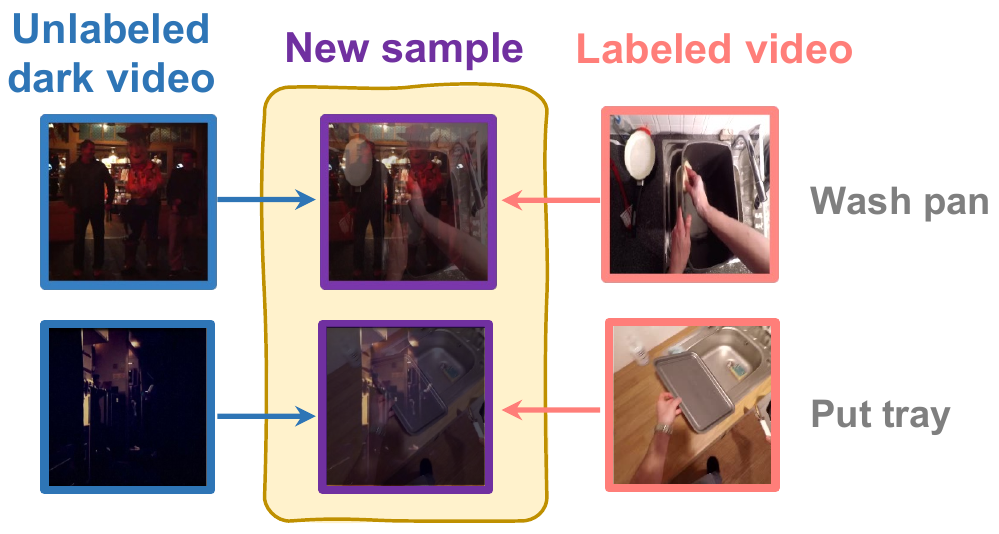}
\vspace{-0.8em}
\caption{
\textbf{Stage 2: Day2Dark-Mix. } Examples of the new videos we create by mixing unlabeled dark videos and labeled day videos. We use these to finetune the model. 
}
\label{fig:mixup}
\end{figure}


\subsection{Stage 2: Finetuning with Day2Dark-Mix}
Our pseudo-supervised learning allows a model to learn useful features for low illumination, 
but has not yet specialized the model to recognize activities in the dark. Inspired by mixup~\citep{zhang2017mixup}, we propose day2dark-mix, where we mix unlabeled, task-irrelevant dark videos and labeled daylight videos. 
Given a randomly sampled labeled video $(x^i, y^i) \in \mathcal{S}$ and a random unlabeled dark video $x^j \in U$, we construct a new sample as: 
\begin{equation}
\begin{split}
    \tilde{x}^{i} &= \alpha x^{i} + (1-\alpha) x^{j},
\end{split}
\end{equation}
where $\alpha \in [0.4,1.0)$ is randomly sampled from a uniform distribution. 
Since our dark videos are unlabeled, different from the original mix-up, we use the label $y^i$ from the daylight video as the label for the new sample in a softmax cross-entropy loss. This gives us `dark' activity videos with which to finetune our model. Examples are shown in Fig.~\ref{fig:mixup}. 

\section{Darkness-Adaptive Activity Recognizer}

While our supervision beyond daylight can train any activity recognizer to generalize to dark environments, the information loss caused by the low color contrast in the dark still remains. As audio is invariant to the illumination changes and contains rich activity information, we propose to incorporate the complementary information within sound along with the vision. 
However, as shown in Fig.~\ref{fig:visual_distribution}, there is a visual distribution shift in low-light conditions so that audio-visual correspondences become harder to establish. Thus, we propose a darkness-adaptive model to best recognize activities in varying illuminations. 

\begin{figure}[t!]
\centering
\includegraphics[width=\linewidth]{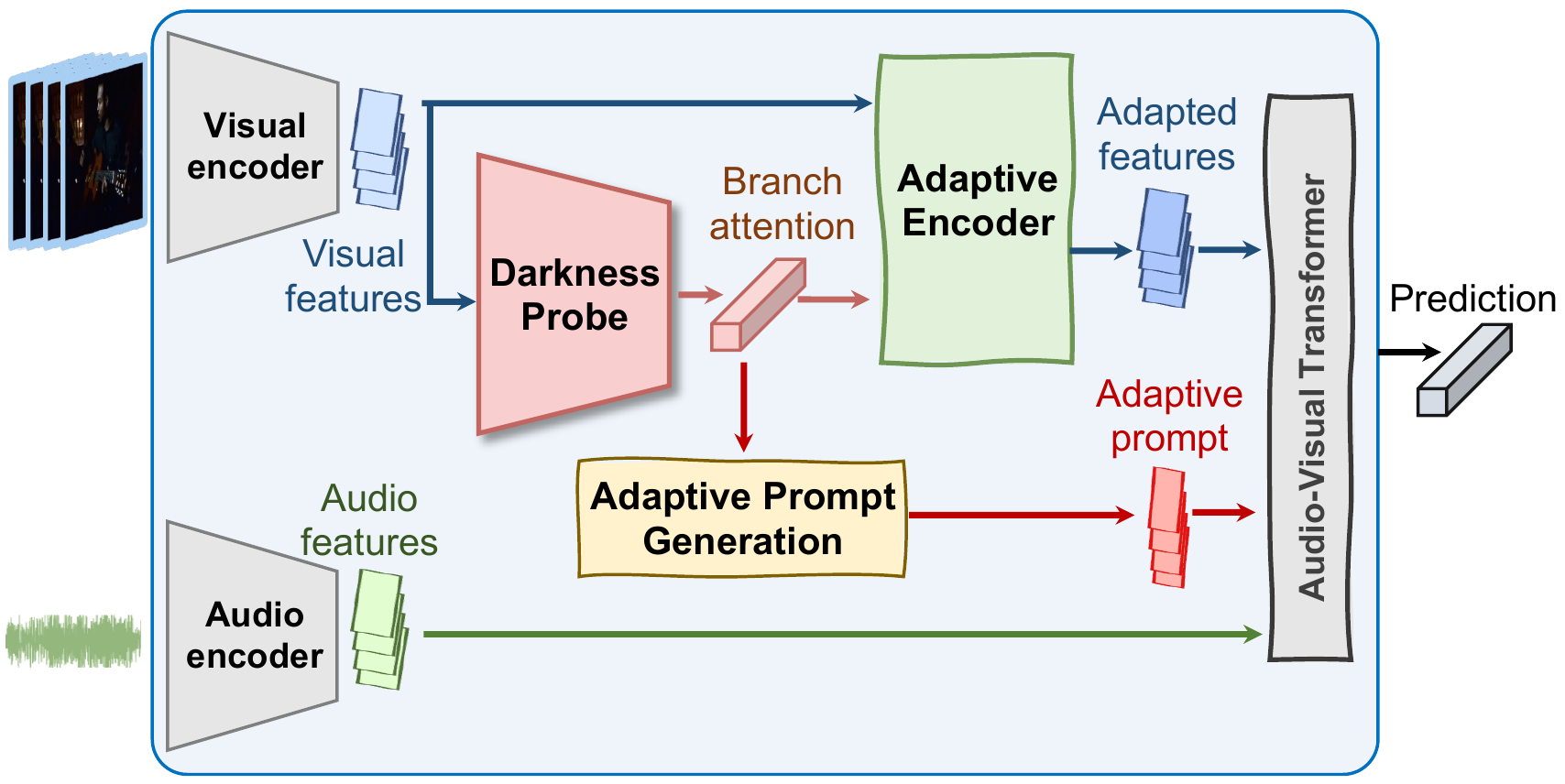}
\vspace{-1.8em}
\caption{\textbf{Darkness-Adaptive Activity Recognizer.} 
\label{sec:adaptive_recognizer}
Given a video, the darkness probe evaluates the clarity of the visual features and outputs a branch attention used by the adaptive encoder and the adaptive prompt generation. The adaptive encoder reduces the visual distribution shift caused by different illuminations. The generated adaptive prompt guides the fusion in the audio-visual transformer according to the illumination. 
Our activity recognizer is trained via supervision beyond daylight (as detailed in Fig.~\ref{fig:supervision}). %
}
\label{fig:framework}
\end{figure}

Our darkness-adaptive model is depicted in Fig.~\ref{fig:framework}. Given a video clip, we extract visual and audio features with single-modality encoders, as in prior works~\citep{lee2021crossattentional,tian2018audio}. 
To compensate for the varying visual feature distribution during the dark, we propose to use $K$ parallel branches in our darkness-adaptive recognizer to handle different visual conditions in the dark. Specifically, the \textit{darkness probe} first analyzes the clarity of the visual features, outputting the branch attention. 
Our \textit{adaptive encoder} uses this attention to reduce the distribution shift in visual features across illuminations by weighting its branches. 
The branch attention also informs the \textit{adaptive prompt generation} which produces an adaptive prompt to guide the audio-visual fusion in the transformer. 
The \textit{audio-visual transformer} takes the adapted visual features, the adaptive prompt and the audio features as inputs, and outputs the activity prediction. 
We detail each component next.

\subsection{Visual and Audio Encoders}
From an input video clip, the visual encoder outputs features $\mathbf{F} {=} \{\mathbf{f}_1, \cdots, \mathbf{f}_{n_v} | \mathbf{f}_i \in \mathbb{R}^{d_v}\}$, which represent different temporal and spatial parts of the video. 
The audio encoder takes the corresponding audio signal as input, and outputs features $\mathbf{A}' {=} \{\mathbf{a}_1, \cdots, \mathbf{a}_{n_a} |  \mathbf{a}_i \in \mathbb{R}^{d_a}\}$ which represent different frequencies and temporal chunks of the audio. 
Since the audio features are independent of illumination, we directly encode them using a single projection layer:
$\mathbf{A} {=} \mathbf{A}'\mathbf{E}^a, \ \mathbf{E}^a \in \mathbb{R}^{d_a \times d_{in}}$, where $d_{in}$ is the input dimension to the audio-visual transformer. 

\subsection{Darkness Probe \& Branch Attention}
We want to adapt the visual features according to video illuminance. We could directly use the estimated illuminance by Eq.\ref{eq:y_compute}. However, this only depicts the overall illuminance, rather than the clarity of the activity. For example, in the `blowing out candles' class, the activity is usually clear because of the light from the candles, even though the overall illuminance is low. Thus, we propose a darkness probe module $\Phi_d$, which has three transformer layers with the same architecture as~\cite{vit}. It takes the features from the visual encoder as input and outputs the branch attention $\bm{\beta} \in \mathbb{R}^{1 \times K}$. This attention is a weighting for the $K$ different branches in the subsequent components that we use to adapt to the visibility of the activity. 

\subsection{Adaptive Encoder}
Our adaptive encoder consists of $K$ parallel projection layers (branches) which encode the visual features according to the darkness. Each projection layer $\mathbf{E}^v_{k} \in \mathbb{R}^{d_v \times d_{in}}$ takes the visual features $\mathbf{F}$ as input and outputs $\mathbf{F}_k {=} \mathbf{F}\mathbf{E}^v_k$. We use our branch attention to integrate the outputs of the $K$ layers by $\mathbf{V} {=} \sum_{k=1}^K \beta_k \mathbf{F}_k$, where $\beta_k$ is the $k$th element in the branch attention $\bm{\beta}$. By using different projections for different input visual distributions, we reduce the distribution shift in the adapted visual features $\mathbf{V} \in \mathbb{R}^{n_v \times d_{in}}$, which are sent to the audio-visual transformer. 

\subsection{Adaptive Prompt Generation}
To recognize activities in the dark, the model should pay more or less attention to motion, appearance and audio according to the light conditions. 
Recent works~\citep{jia2022visual,promptcontinual,zhou2022learning} show different learnable prompts allow a transformer to focus on different tasks. Inspired by this, we treat different levels of darkness as different tasks and learn $K$ prompts, $\{\mathbf{O}_1, \cdots, \mathbf{O}_K | \mathbf{O}_i \in \mathbb{R}^{l \times d_{in}}\}$, where each prompt consists of $l$ tokens. We obtain the illuminance-adaptive prompt by $\mathbf{O} {=} \sum_{k=1}^K \beta_k \mathbf{O}_k$, which aggregates the $K$ prompts with our branch attention.

\subsection{Audio-Visual Transformer}
The encoded visual features, audio features and adaptive prompt are concatenated as $[\mathbf{V}, \mathbf{A}, \mathbf{O}]$ and sent to a transformer for audio-visual fusion. The output is also passed to $K$ parallel classifiers $g_1(\cdot), \cdots, g_K(\cdot)$ for different light conditions. Similarly to the prompt, the final classification result is obtained from a weighting with branch attention $\beta$: $\mathbf{y} {=} \sum_{k=1}^K \beta_k \mathbf{y}_k$, where $\mathbf{y}_k \in \mathbb{R}^{1\times n}$ is the prediction from classifier $g_k(\cdot)$ for $n$ classes. We treat the number of branches as a hyperparameter and use $K{=}5$ branches by default. For daylight videos, \ie, $Y{>}t$, we find a single branch, consisting of the projection layer, prompts and classification head suffices for the fusion since the visual features are clear and their distribution remains similar across illuminations.

\subsection{Training with Supervision beyond Daylight}
As explained in Section~\ref{sec:supervision}, our supervision beyond daylight trains an activity recognizer in two stages: 

\noindent\textit{Stage 1:} Our darkness-adaptive activity recognizer is trained with our pseudo-supervised day2dark learning to learn to fuse audio and visual information for the auxiliary tasks in the dark. 

\noindent\textit{Stage 2:} The first stage allows the model to learn a useful representation to distinguish dark videos, however it is not yet specialized for the target task of activity recognition. Specifically, we freeze the audio-visual transformer in our activity recognizer and use our day2dark-mix to finetune the darkness-adaptive components: the darkness probe, the adaptive encoder and the adaptive prompt generation, for the target task only. 

By making our activity recognizer `darkness-adaptive' and training it with supervision beyond daylight, we can considerably reduce the day2dark gap observed in Section~\ref{sec:day2darkgap}. 

\begin{figure}[t!]
\centering 
\includegraphics[width=\linewidth]{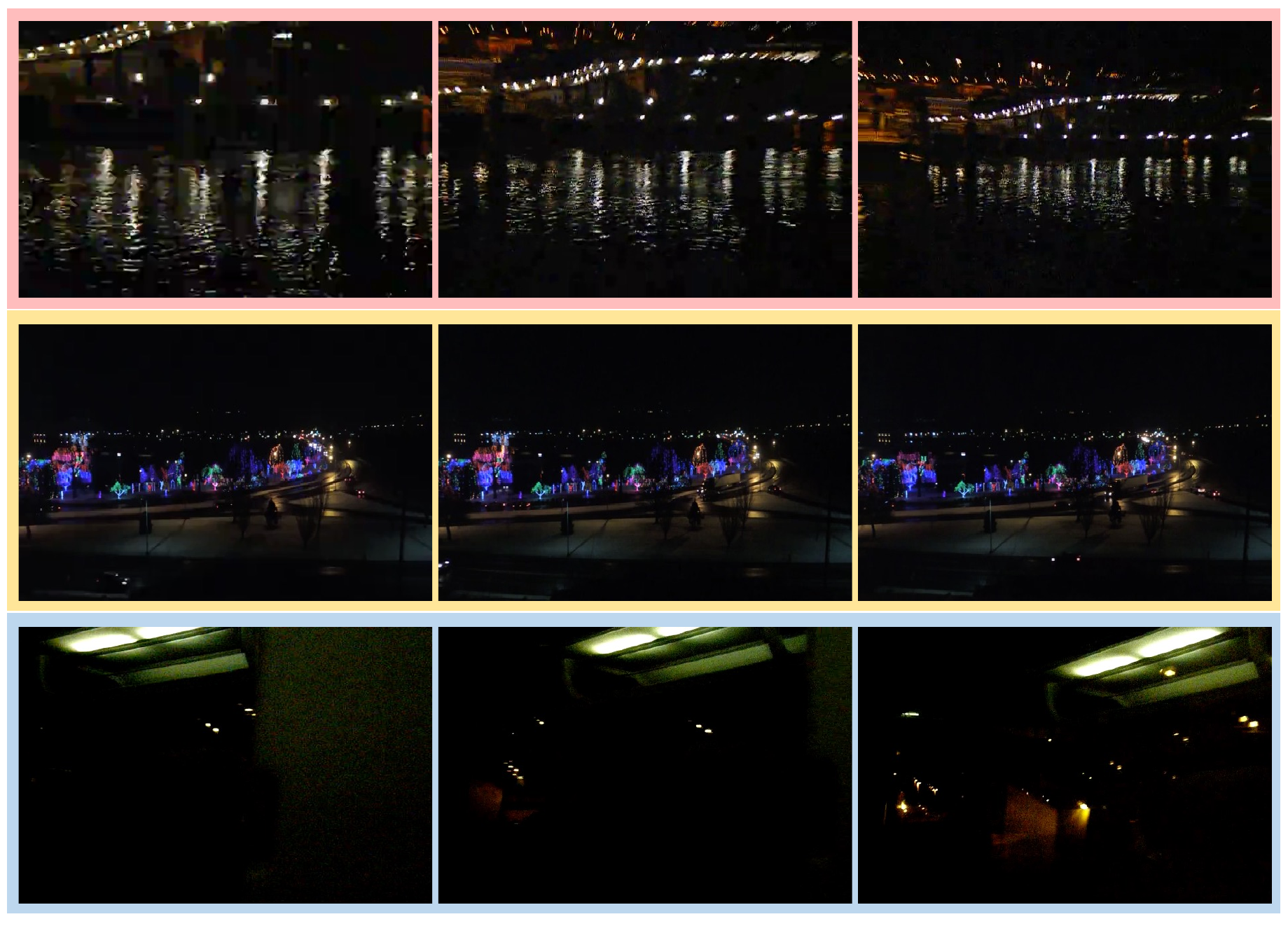}
\vspace{-2em}
\caption{
\textbf{Examples of Unlabeled Dark Videos.} 
Each row contains three random frames of a single video. 
}
\label{fig:examples_unlabeled}
\end{figure}

\section{Experiments}
\label{sec:experiments}

We first describe the adopted datasets, implementation details and evaluation criteria before ablating the components of our method and comparing it to prior works for image enhancement, domain adaptation and audio-visual fusion.  

\begin{table*}[t!]
\centering
\resizebox{0.85\linewidth}{!}{
    \begin{tabular}{lccccccc}
    \toprule
    & \multicolumn{2}{c}{\textbf{EPIC-Kitchens}} & \multicolumn{2}{c}{\textbf{Kinetics-Sound}} & \multicolumn{2}{c}{\textbf{Charades}}\\
    \cmidrule(lr){2-3} \cmidrule(lr){4-5} \cmidrule(lr){6-7}
\textbf{Model} & Dark $\uparrow$ & Day $\uparrow$ & Dark $\uparrow$ & Day $\uparrow$ & Dark $\downarrow$ & Day $\downarrow$ \\
\midrule
\rowcolor{audiogreen}
\textbf{Audio only} & & & & & &\\
Audio encoder & 12.8 & 10.8 & 51.5 & 59.3 & 0.585 & 0.554 \\
\rowcolor{visualblue}
\textbf{Visual only} & & & & & &\\
Visual encoder & 26.6 & 37.0 & 77.2 & 82.2 & 0.047 & 0.026 \\
+ Pseudo-supervised day2dark learning & 27.4 & 37.1 & 78.5 & 83.5 & 0.043 & 0.025 \\
+ Day2dark-mix finetuning & 27.9 & 37.2 & 79.2 & 83.9 & 0.041 & 0.025 \\
\rowcolor{gray!20}
\textbf{Audio \& Visual} & & & & & &\\
Darkness-adaptive activity recognizer & 29.3 & 37.4 & 80.3 & 85.7 & 0.038 & 0.024 \\
+ Pseudo-supervised day2dark learning & 34.2 & 37.5 & 84.2 & 86.7 & 0.031 & 0.023 \\
+ Day2dark-mix finetuning & 35.6 & 37.6 & 85.1 & 87.2 & 0.028 & 0.022 \\
\bottomrule
\end{tabular}
}
\vspace{-0.5em}
\caption{\textbf{Ablation of Supervision Beyond Daylight.} Pseudo-supervised day2dark learning and day2dark-mix increase accuracy of both a visual-only and our darkness-adaptive activity recognizer in the dark and maintain performance in the day. %
}
\label{tab:supervision_beyond_daylight_ablation}
\end{table*}

\subsection{Datasets}
\vspace{-0.3em}
\label{sec:datasets}
\noindent \textbf{Unlabeled Dark Videos.} 
We source unlabeled dark videos from YFCC100M~\citep{thomee2016yfcc100m} due to its scale. %
From Section~\ref{sec:day2darkgap}, we observe dark videos commonly satisfy $Y{\leq}40$ (Eq.~\ref{eq:y_compute}), thus we set our illuminance threshold to $t{=}40$. 
We treat videos with $Y{\leq}t$ as `dark' and videos with $Y{>}t$ as `day'. 
We ensure our model works in the extreme case, where all the unlabelled dark videos are task-irrelevant, by removing any video our trained model classifies as any activity with confidence ${>}0.5$. 
With this constraint, we obtain 8,800 dark videos.
We manually check these videos and find that they do not contain activities we aim to recognize and usually contain no human activities at all. Three typical examples are provided in Fig.~\ref{fig:examples_unlabeled}. 
%


\noindent \textbf{EPIC-Kitchens.} EPIC-Kitchens-100~\citep{EPIC} contains first-person videos. Training and validation sets include 67,217 and 9,668 video clips, each annotated with a verb and a noun, from 97 verbs and 300 nouns. There are 1,053 and 372 dark clips in training and validation. 


\noindent \textbf{Kinetics-Sound.} Kinetics-Sound~\citep{KineticsSound} is a subset of Kinetics-400~\citep{Kinetics} with 34 activities. %
The training, validation and test sets contain 23,847, 1,652 and 2,691 videos, where 1,927, 123 and 303 are dark. %

\noindent \textbf{Charades. }
Charades~\citep{sigurdsson2016hollywood} contains 9,848 third-person videos with 66,500 multi-label annotations for 157 daily, indoor activities. Train and test consist of 7,985 and 1,863 videos, where 274 and 82 are dark. 

\vspace{-0.2em}
\subsection{Implementation \& Evaluation Details}
\vspace{-0.3em}


\noindent \textbf{Supervision Beyond Daylight.} 
We use two auxiliary tasks with off-the-shelf models~\citep{gabeur2020multi,mo2022localizing}: video-text matching on HowTo100M~\citep{miech2019howto100m} and self-supervised sound source localization on VGGSound~\citep{VGGSound}. 
The video-text matching pseudo-label is the visual feature embedding, we do not use text. 
For sound source localization we use the activation map. 

\noindent \textbf{Darkness-Adaptive Activity Recognizer.} For the audio encoder, we adopt ResNet-18~\citep{resnet}, pre-trained on \mbox{VGGSound}~\citep{VGGSound}. 
We test our model with two different visual encoders. Unless otherwise specified we use Swin-T~\citep{liu2021swin}, pre-trained with Omnivore~\citep{girdhar2022omnivore}, as default. When reporting results on SlowFast~\citep{slowfast}, the backbone has been pretrained with Kinetics-400~\citep{Kinetics}. 
Input video clips contain 32 frames with a width and height of 224. The darkness probe and audio-visual transformer consist of 3 and 6 transformer layers~\citep{vit}, with 8 heads and a hidden size of 256. We use $K{=}5$ as the branch attention dimension and number of projection layers, prompts and classifiers. 
Each adaptive prompt contains 10 tokens. 
The first output token of the prompt is used as the activity prediction and the second for the pseudo-label in our pseudo-supervised learning.

\noindent \textbf{Training.} We use SGD with a momentum of 0.9 and a batch size of 32. As explained in Section~\ref{sec:supervision}, the training contains two stages, we use a learning rate of 0.01 in stage 1 and 0.3 in stage 2. We share all our code on the project page: \url{https://xiaobai1217.github.io/Day2Dark/}.

\noindent \textbf{Evaluation Criteria. }
Following standard practice \citep{EPIC,KineticsSound}, we report top-1 accuracy for EPIC-Kitchens and Kinetics-Sound. Since mean average precision (mAP) is dependent on test-set size, we report the Hamming distance on day and dark test sets for multi-label classification on Charades.

\subsection{Supervision Beyond Daylight}

\noindent \textbf{Components Ablation.} 
We show the ablation in Table~\ref{tab:supervision_beyond_daylight_ablation}. When training a visual-only model with our pseudo-supervised learning and day2dark-mix finetuning, we obtain improvements of 1.3\% and 0.7\% in the dark on Kinetics-Sound, while maintaining good performance in the day. 
EPIC-Kitchens and Charades have similar improvements. 
Thus, using unlabeled dark videos improves the generalizability of the visual encoder towards low illumination, even though these videos are task-irrelevant. Table~\ref{tab:supervision_beyond_daylight_ablation} also highlights that our supervision beyond daylight is key to training our darkness-adaptive recognizer. 
Specifically, on EPIC-Kitchens and Kinetics-Sound, the accuracy increases by 4.9\% in the dark with our pseudo-supervised learning, as this allows the model to find better audio-visual correspondences in low illumination. 
Finetuning with day2dark-mix results in further 1.4\% and 0.9\% improvements since the model becomes better suited for the target task in the dark. 
We also see considerable decreases in Hamming distance on Charades in the dark. 
For all datasets, adding our supervision beyond daylight does not degrade performance in the day, while performance in the dark increases considerably.

\begin{figure}[t!]
\centering 
\includegraphics[width=0.7\linewidth]{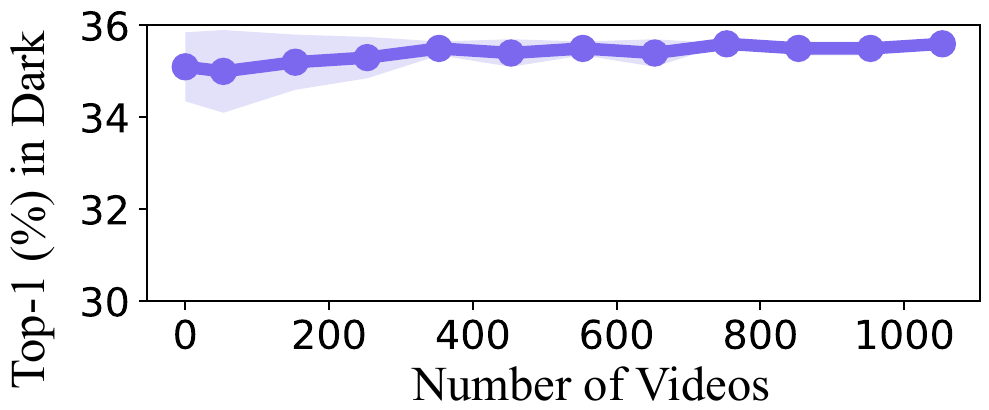}
\vspace{-0.5em}
\caption{\textbf{Number of Labeled Dark Videos} on EPIC-Kitchens. Our model is successful even when there are no labeled dark videos in training. 
}
\label{fig:number_of_labeled}
\end{figure}

\begin{figure}[t!]
\centering 
\includegraphics[width=0.7\linewidth]{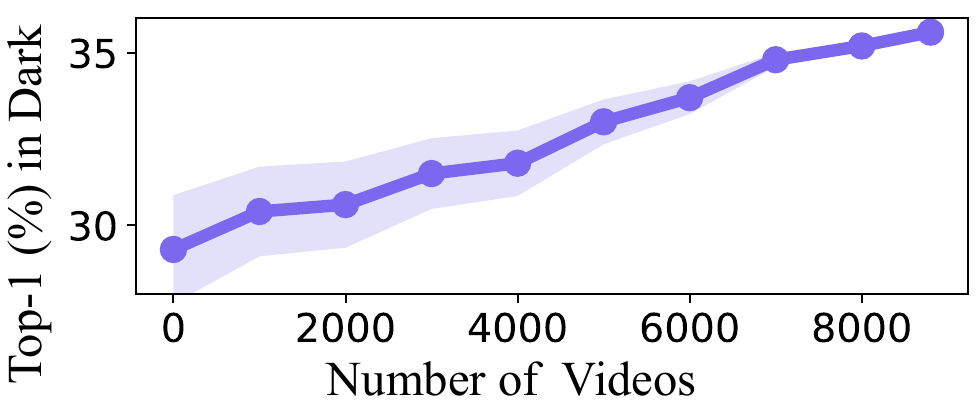}
\vspace{-0.5em}
\caption{\textbf{Number of Unlabeled Dark Videos} on EPIC-Kitchens. Increasing the number of unlabeled dark videos improves accuracy in the dark.
}
\label{fig:number_of_unlabeled}
\end{figure}

\noindent \textbf{Effect of Number of Labeled Dark Videos.} 
To demonstrate our model works in the extreme case without labeled dark videos in training, we vary the number of labeled dark videos used for EPIC-Kitchens in Fig.~\ref{fig:number_of_labeled}. We see that removing all labeled dark videos for training has little impact on performance and delivers 35.1\% \textit{vs.} 35.6\% accuracy. 
For different numbers of labeled dark videos, the accuracy variance is merely 0.035\%. 
Thus, our supervision beyond daylight can effectively utilize task-irrelevant dark videos, circumventing the need for dark videos with activity labels during training.

\noindent \textbf{Effect of Number of Unlabeled Dark Videos. }
In Fig.~\ref{fig:number_of_unlabeled} we investigate the effect of the number of unlabeled dark videos in training with EPIC-Kitchens. These results show that our supervision beyond daylight is more effective when using more unlabeled dark videos. %
This is because they provide a rich dark data distribution for model learning and reduce overfitting to daylight videos.

\noindent \textbf{Effect of Auxiliary Tasks. }
Table~\ref{tab:source_of_pseudo} ablates the effectiveness of the two auxiliary tasks used in our pseudo-supervised day2dark learning: video-text matching and sound source localization. 
Using one task gives up to +3.4\% improvement in the dark while the combination results in +5.5\%. 
This is because multiple tasks contain a wider label space, improving our model's ability to generalize to the target task.

\begin{table}[t!]
\centering
\resizebox{\linewidth}{!}{
    \begin{tabular}{cccccccccccc}
    \toprule
\multicolumn{2}{c}{\textbf{Auxiliary Task}} &  
\multicolumn{2}{c}{\textbf{EPIC-Kitchens}}\\
\cmidrule(lr){1-2} \cmidrule(lr){3-4}
Video-Text & Sound Source & \\
 Matching &  Localization & Dark $\uparrow$ & Day $\uparrow$ \\
 \citep{miech2019howto100m} & \citep{VGGSound} \\
        \midrule
 & & 30.1 & 37.5 \\
\CheckmarkBold & & 33.3 & 37.5 \\
 & \CheckmarkBold & 33.5 & 37.6 \\
\CheckmarkBold & \CheckmarkBold & 35.6 & 37.6 \\
    \bottomrule
    \end{tabular}
}
\vspace{-0.6em}
\caption{\textbf{Effect of the Auxiliary Tasks} in pseudo-supervised day2dark learning. Using either task improves performance in the dark and their combination gives further improvement. 
}
\label{tab:source_of_pseudo}
\end{table}

\begin{figure*}[htp]
\includegraphics[width=\linewidth,height=0.34\linewidth]{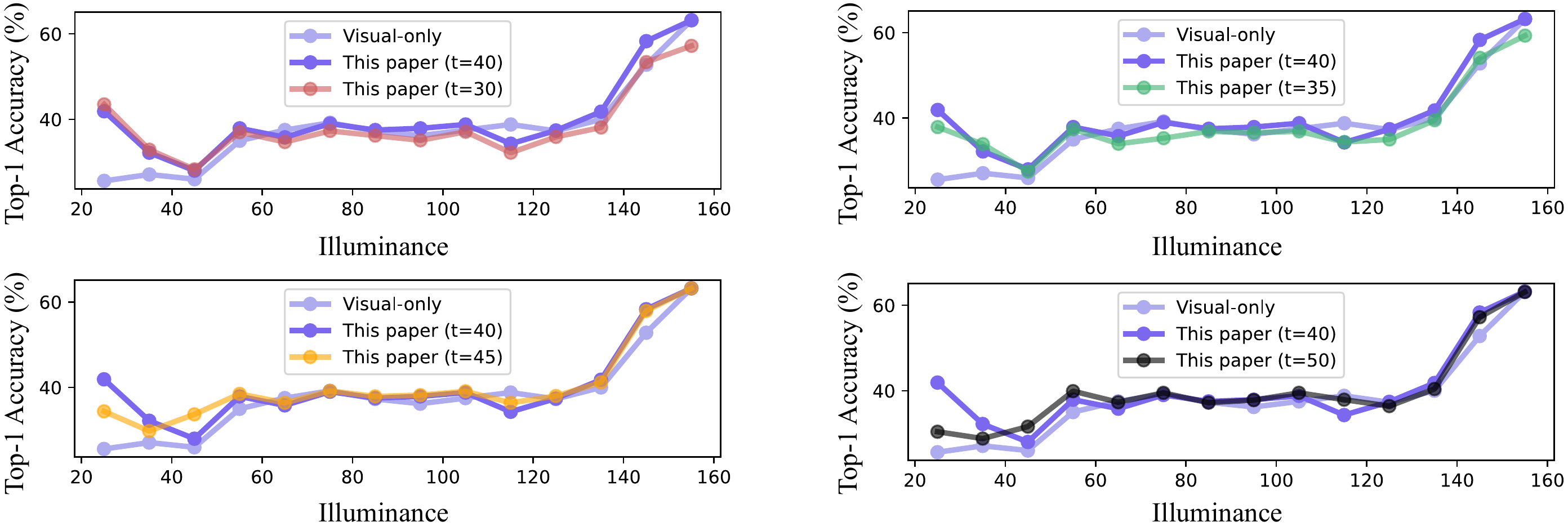}
\vspace{-1.8em}
\caption{\textbf{Effect of threshold $t$} to obtain unlabeled dark videos from YFCC100M~\citep{thomee2016yfcc100m} using illuminance $Y {\leq} t$ for selection. For all values of $t$ tested, our model reduces the day2dark gap considerably. 
A smaller $t$ focuses the model on the lowest illuminances, while a large $t$ is less effective on these low illuminances. We find threshold $t{=}40$ delivers the best trade-off. }
\label{fig:threshold_t}
\end{figure*}

\noindent \textbf{Source of Pseudo Labels.} 
Instead of using predictions of auxiliary tasks for pseudo-supervised day2dark learning, here we consider using the predictions of the target task directly. 
Specifically, we first get the predictions from the trained visual-only activity classification model for the unlabeled dark videos. 
Then we use these predictions to train an autoencoder and obtain the abstract representations as the pseudo labels for Eq.~\ref{eq:L_U}. 
Using such pseudo labels, we obtain 30.4\% in the dark, compared to our 33.3\% with one auxiliary task and 35.6\% with two auxiliary tasks. The performance in the daylight also degrades to 37.1\%, while ours is 37.6\%. Because the unlabeled videos do not contain activities of interest, meaning such pseudo-labels are uninformative and cannot provide a useful training signal.

\noindent \textbf{Abstract Representation \textit{vs.} Predictions. }
In our pseudo-supervised day2dark learning, we use an autoencoder to encode the pseudo labels of auxiliary tasks. The latent bottleneck representation from the autoencoder is used as the final pseudo label for training the activity recognizer. We could instead directly use the auxiliary task predictions as the pseudo labels. However, this is much less effective giving 31.5\% and 37.4\% in the dark and day on EPIC-Kitchens, worse than our bottleneck representation which gives 35.6\% and 37.6\%. This is because directly using the pseudo labels, which could have a high dimension (\eg localization maps for sound source localization), can cause the model to overfit due to the additional parameters introduced for fitting the pseudo labels. In contrast, the bottleneck representation is more compact and thus more abstract, reducing overfitting. We conclude that using a compact representation is preferred over the original predictions for the auxiliary tasks.

\noindent \textbf{Effect of $t$ for Collecting Unlabeled Dark Videos. }
Our supervision beyond daylight needs unlabeled dark videos for model learning. 
In Section~\ref{sec:datasets}, we collect videos with illuminance $Y{\leq}t$ from YFCC100M~\citep{thomee2016yfcc100m} into our set of unlabeled dark videos and we set the threshold $t=40$ by default. We ablate the effect of threshold $t$ in Fig.~\ref{fig:threshold_t}. Our model reduces the day2dark gap for all values of $t$ tested. However, with a lower threshold ($t{\leq}35$), our model struggles in daylight scenarios, as it over-fits to the lowest illuminances. In contrast, with a larger threshold ($t{\geq}45$), our model becomes less effective on the lowest light conditions than with $t{\leq}40$. When setting threshold $t{=}40$, we obtain the best trade-off. 


\begin{table}
\centering
\resizebox{0.5\linewidth}{!}{
\begin{tabular}{lcc}
\toprule
& \multicolumn{2}{c}{\textbf{EPIC-Kitchens}} \\
\cmidrule(lr){2-3}
$\lambda$ & Dark $\uparrow$ & Day $\uparrow$ \\
\midrule
0.008 & 34.9 & 37.5 \\
0.009 & 35.3 & 37.6 \\
0.01 & 35.6 & 37.6 \\
0.02 & 35.4 & 37.2 \\
0.03 & 35.2 & 37.0 \\
\bottomrule
\end{tabular}
}
\vspace{-0.5em}
\caption{\textbf{Effect of $\lambda$} to balance the loss terms in Eq.~\ref{eq:pseudo_supervised}. A larger $\lambda$ results in overfitting to the auxiliary tasks, while a smaller $\lambda$ cannot enable the model to effectively learn from the unlabeled dark videos. With $\lambda=0.01$, we obtain the best trade-off. }
\label{tab:effect_of_lambda}
\end{table}


\noindent \textbf{Effect of $\lambda$ to Balance the Loss Terms.} In Eq.~\ref{eq:pseudo_supervised}, we set $\lambda=0.01$ to balance the losses between pseudo-supervised learning and classification learning of the target task. Here, we ablate its effect in Table~\ref{tab:effect_of_lambda}. Our model is robust to the choice of $\lambda$ and we find $\lambda=0.01$ gives the best balance between learning from day and dark videos.

\begin{table*}[t!]
\centering
\resizebox{0.85\linewidth}{!}{
    \begin{tabular}{lcccccc}
    \toprule
    & \multicolumn{2}{c}{\textbf{EPIC-Kitchens}} & \multicolumn{2}{c}{\textbf{Kinetics-Sound}} & \multicolumn{2}{c}{\textbf{Charades}} \\
    \cmidrule(lr){2-3} \cmidrule(lr){4-5} \cmidrule(lr){6-7}
\textbf{Model} & Dark $\uparrow$ & Day $\uparrow$ & Dark $\uparrow$ & Day $\uparrow$ & Dark $\downarrow$ & Day $\downarrow$ \\
\midrule
\rowcolor{gray!20}
\textbf{Audio \& Visual} & & & & & & \\
Vanilla multi-modal transformer & 29.8 & 37.2 & 81.2 & 86.7 & 0.039 & 0.023 \\
+ Darkness probe \& adaptive encoder & 31.8 & 37.4 & 82.3 & 86.8 & 0.036 & 0.023 \\
+ Adaptive prompts & 34.3 & 37.5 & 83.9 & 87.1 & 0.030 & 0.022 \\
+ Adaptive classification  & 35.6 & 37.6 & 85.1 & 87.2 & 0.028 & 0.022 \\
\bottomrule
\end{tabular}
}
\vspace{-0.5em}
\caption{\textbf{Ablation of Darkness-Adaptive Activity Recognizer.} All model variants are trained with supervision beyond daylight. 
Each component increases performance in the dark, without decreasing the accuracy in daylight conditions. 
}
\label{tab:darkness_fusion_ablation}
\end{table*}

\begin{figure*}[t!]
\centering 
\includegraphics[width=1\linewidth,height=0.33\linewidth]{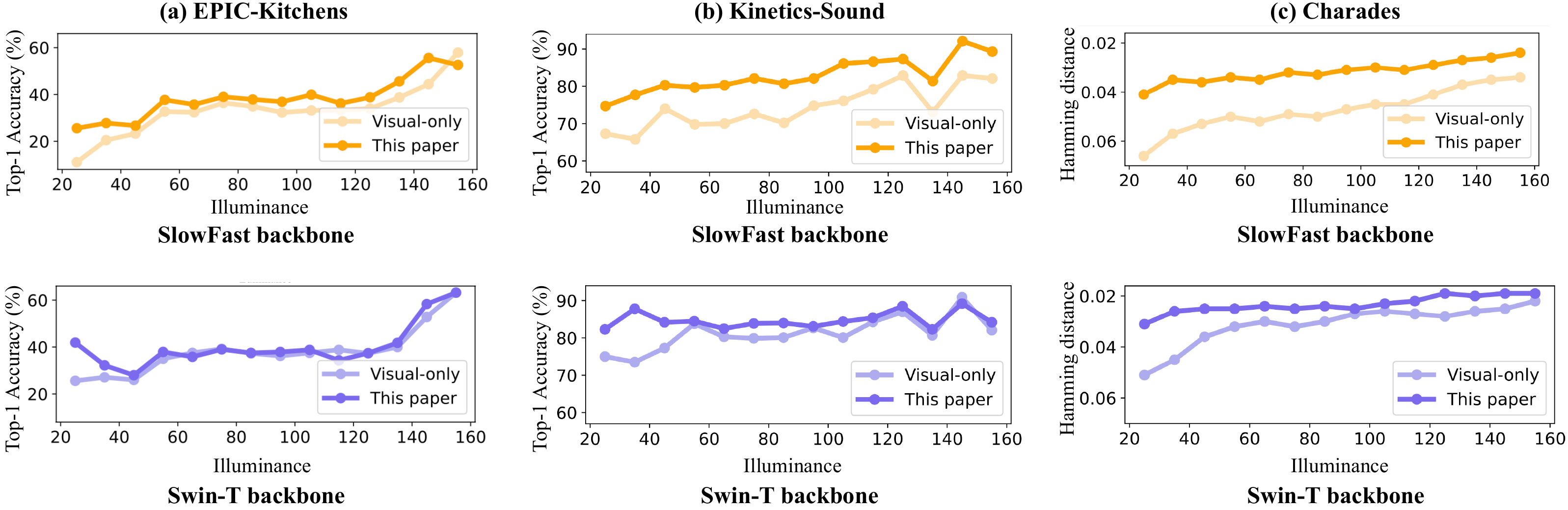}
\vspace{-1.5em}
\caption{\textbf{Benefit Over Different Illuminances for (a) EPIC-Kitchens, (b) Kinetics-Sound and (c) Charades} with the SlowFast and Swin-T backbones. 
Our model improves performance in both the day and dark. The effectiveness of our model increases in the dark, successfully reducing the day2dark gap for both datasets.
}
\label{fig:benefit_illuminance}
\end{figure*}

\subsection{Darkness-Adaptive Activity Recognizer}

\noindent \textbf{Components Ablation.} 
We present the ablation in Table~\ref{tab:darkness_fusion_ablation}. 
Audio contains rich activity information. 
However, fusing audio and visual features by a vanilla multi-modal transformer~\citep{gabeur2020multi} gives only a small improvement, \eg, +1.9\% on EPIC-Kitchens, since the visual features are less reliable for fusion in the dark. Adding our adaptive encoder to reduce the visual distribution shift gives a 2.0\% increase on EPIC-Kitchens and the adaptive prompt contributes a further 2.5\% as it stresses different aspects of each modality according to illumination. On Kinetics-Sound and Charades we obtain similar improvements. The combination of adaptive encoder, prompts, and classification heads delivers the best results, since each classification head focuses on different features that the adaptive prompt stresses. 
Our darkness-adaptive fusion can even improve activities without characteristic sound, \eg, `reading a book' and `holding some food' in Charades, and also gives a small improvement (+0.4\% and +0.1\% on EPIC-Kitchens and Kinetics-Sound) in the day. 
To verify that our darkness probe is needed, we replace it with the illuminance computed by Eq.~\ref{eq:y_compute}. This degrades performance by -2.1\%, -2.0\%, and +0.004 (Hamming distance) in the dark for EPIC-Kitchens, Kinetics-Sound and Charades. Thus, making our model `darkness-adaptive' allows for better audio-visual fusion under different illuminations and reduces the day2dark gap considerably. 

We provide further ablations for both the supervision beyond daylight and dark-adaptive activity recognizer with the convolutional SlowFast backbone in the Appendix. Our method also provides a more effective cross-modal fusion in the dark with a convolutional backbone.

\begin{table}
\centering
\resizebox{0.45\linewidth}{!}{
\begin{tabular}{lcc}
\toprule
& \multicolumn{2}{c}{\textbf{EPIC-Kitchens}} \\
\cmidrule(lr){2-3}
$K$ & Dark $\uparrow$ & Day $\uparrow$ \\
\midrule
1 & 29.8 & 37.2 \\
2 & 30.4 & 37.2 \\
3 & 32.5 & 37.4 \\
4 & 34.3 & 37.4 \\
5 & 35.6 & 37.6 \\
6 & 34.8 & 37.5 \\
7 & 33.5 & 37.3 \\
\bottomrule
\end{tabular}
}
\vspace{-0.5em}
\caption{\textbf{Effect of $K$}. Using multiple branches is always better than a single branch ($K=1$). Overall, $K{=}5$ gives the best trade-off between under- and over-fitting. }
\label{tab:effect_of_k}
\end{table}


\noindent \textbf{Effect of $K$}. 
In our darkness-adaptive activity recognizer, we use $K{=}5$ branches of projection layers, prompts and classification heads by default. 
Here we ablate the effect of $K$ in Table~\ref{tab:effect_of_k}. 
We find that using multiple branches is always better than a single branch. 
While fewer branches can result in underfitting, more branches can cause overfitting. 
Overall, $K{=}5$ provides the best trade-off.

\subsection{Analysis}

\noindent \textbf{Benefit Over Different Illuminances.} 
In Fig.~\ref{fig:benefit_illuminance}, we show the benefit of our method over the visual-only encoder for different illuminances on EPIC-Kitchens, Kinetics-Sound and Charades. 
We present the results with both the SlowFast and the Swin-T backbone. 
As videos become darker, our model increases in effectiveness. 
For example, on EPIC-Kitchens, with illuminance $30{<}Y{\leq}40$, our model with Swin-T backbone gives a +5.1\% improvement, while with $20{<}Y{\leq}30$, our model gives a larger increase of +16.3\%. 
On Kinetics-Sound and Charades, we also observe larger performance improvements with $Y{\leq}40$ than good light conditions with $Y{\geq}40$. 
With the SlowFast backbone, while there are larger accuracy increases on EPIC-Kitchens in the dark than in the day, our method consistently improves over the visual-only encoder across all illuminances on Kinetics-Sound and Charades owing to the incorporation of the complementary information within sound. 
Thus, we conclude that our method is beneficial for either transformer or convolutional backbones under low illuminations while maintaining high accuracy in the daylight. 
%
%

\begin{table*}
\centering
\resizebox{0.65\linewidth}{!}{
\begin{tabular}{lcr}
\toprule
& \multicolumn{2}{c}{\textbf{EPIC-Kitchens}} \\
\cmidrule(lr){2-3}
\textbf{Model} & Dark $\uparrow$ & GFLOPs $\downarrow$ \\
\midrule
\rowcolor{Gray}
\textbf{Baseline} & &\\
Vanilla multi-modal transformer & 29.8 & 1.4 \\
\rowcolor{Gray}
\textbf{Image enhancement} & & \\
KinD~\citep{zhang2019kindling} & 20.3 & 932.2 \\ 
SCI \citep{ma2022toward} & 24.1 & 3.4\\ 
KinD-plus~\citep{zhang2021beyond} & 24.6 & 1321.6 \\ 
URetinex-Net~\citep{wu2022uretinex} & 24.7 & 152.0 \\ 
GLADNet \citep{wang2018gladnet} & 25.5 & 8068.5 \\ 
RUAS~\citep{liu2021retinex} & 25.9 & 10.4  \\ 
Unsupervised enhancement \citep{jin2022unsupervised} & 26.4 & 108.8 \\ 
LCDPNet \citep{wang2022enhancement} & 26.5 & 422.4 \\ 
LEDNet \citep{zhou2022lednet} & 27.8 & 312.0 \\ 
Gamma Intensity Correction~\citep{poynton2012digital}
& 29.7 & 22.7 \\ 
\rowcolor{Gray}
\textbf{Day2Dark model} & & \\%
\textit{\textbf{This paper}} & 35.6 & 1.6 \\
\bottomrule
\end{tabular}
}
\vspace{-1em}
\renewcommand\thetable{8}
\caption{\textbf{Comparison with image enhancement} on EPIC-Kitchens. We report the FLOPs needed to process a single 32 frame clip alongside accuracy in the dark. Our approach is superior to image enhancement techniques in the dark, which degrade the performance of a vanilla multi-modal transformer through the distortions they introduce while increasing computation time. 
}
\label{tab:image_enhancement}
\end{table*}


\begin{table}
\centering
\resizebox{0.9\linewidth}{!}{
\begin{tabular}{lcc}
\toprule
& \multicolumn{2}{c}{\textbf{EPIC-Kitchens}} \\
\cmidrule(lr){2-3}
\textbf{Model} & Dark $\uparrow$ & Day $\uparrow$\\
\midrule
\rowcolor{Gray}
Visual-only & & \\
Visual encoder & 26.6 & 37.0 \\
w/ color jittering & 26.8 & 37.2 \\
w/ illumination adjustment & 27.0 & 37.0 \\
\rowcolor{Gray}
Audio-Visual & & \\
Darkness-adaptive recognizer & 29.3 & 37.4 \\
w/ color jittering & 31.3 & 37.4 \\
w/ illumination adjustment & 31.8 & 37.3 \\
\textit{\textbf{This paper}} & 35.6 & 37.6 \\
\bottomrule
\end{tabular}
}
\vspace{-0.5em}
\renewcommand\thetable{6}
\caption{\textbf{Comparison with Data Augmentation}. Our supervision beyond daylight that utilizes widely unlabeled dark videos is more effective than simple data augmentations. }
\vspace{-1em}
\label{tab:augmentation}
\end{table}


\noindent \textbf{Comparison with Data Augmentations}. 
We compare our day2dark-mix with strong color jittering and illumination augmentations in Table~\ref{tab:augmentation}. 
Using these augmentations on the training data to train the same visual-only encoder results in limited performance improvements in the dark. 
When using them to train our darkness-adaptive recognizer, the performance is still worse than training with our pseudo-supervised day2dark learning and day2dark-mix, which achieves 35.6\%/37.6\% since the augmented videos have a different distribution to real dark videos, which have small well-lit regions from fires, lamps, tv, \etc. %
Utilizing easy to obtain unlabeled dark videos is more effective than simple data augmentations.

\begin{table}
\centering
\resizebox{0.95\linewidth}{!}{
\begin{tabular}{lc}
\toprule
& \textbf{EPIC-Kitchens} \\
\cmidrule(lr){2-2}
\textbf{Model} & Top-1 (\%) $\uparrow$  \\
\midrule
Visual-only & 26.4 \\
Vanilla multi-modal transformer & 27.7 \\
\textit{\textbf{This paper}} & 31.1 \\
\bottomrule
\end{tabular}
}
\vspace{-0.5em}
\renewcommand\thetable{7}
\caption{\textbf{Benefit for Occlusions}. Our darkness-adaptive model is more robust to occlusions since it adapts the audio-visual fusion based on how clear the activity is.}
\label{tab:occlusion}
\end{table}


\begin{table*}
\centering
\resizebox{0.6\linewidth}{!}{ 
\setlength{\tabcolsep}{3pt}
\begin{tabular}{lcccccc}
\toprule
& \multicolumn{2}{c}{\textbf{EPIC-}} & \multicolumn{2}{c}{\textbf{Kinetics}}\\
& \multicolumn{2}{c}{\textbf{Kitchens}} & \multicolumn{2}{c}{\textbf{Sound}} & \multicolumn{2}{c}{\textbf{Charades}} \\
\cmidrule(lr){2-3} \cmidrule(lr){4-5} \cmidrule(lr){6-7} 
\textbf{Model} & Dark $\uparrow$ & Day $\uparrow$ & Dark $\uparrow$ & Day $\uparrow$ & Dark $\downarrow$ & Day $\downarrow$ \\
\midrule
\rowcolor{audiogreen}
\textbf{Audio only} & & & & & &\\
Audio-only & 12.8 & 10.8 & 51.5 & 59.3 & 0.585 & 0.554 \\
\rowcolor{visualblue}
\textbf{Visual only} & & & & & &\\
Visual-only & 26.6 & 37.0 & 77.2 & 82.2 & 0.047 & 0.026 \\
\rowcolor{gray!20}
\textbf{Audio \& Visual} & & & & & &\\
\cite{tian2020unified} & 27.4 & 37.3 & 78.1 & 83.3 & 0.044 & 0.024 \\
\cite{lee2021crossattentional} & 28.3 & 37.5 & 79.8 & 84.1 & 0.039 & 0.024 \\
\cite{gabeur2020multi} & 28.5 & 37.4 & 79.6 & 84.4 & 0.042 & 0.023 \\
\cite{nagrani2021attention} & 28.6 & 37.5 & 79.9 & 84.7 & 0.037 & 0.022 \\
\textbf{\textit{This paper}} & 35.6 & 37.6 & 85.1 & 87.2 & 0.028 & 0.022 \\
\bottomrule
\end{tabular}
}
\vspace{-0.5em}
\renewcommand\thetable{9}
\caption{\textbf{Comparison with Audio-Visual Fusion}. 
Although audio-visual methods generalize better than visual-only, our method provides a more effective cross-modal fusion in the dark. 
}
\label{tab:alternative_fusion}
\end{table*}

\noindent \textbf{Benefit For Occlusions. }
In Table~\ref{tab:occlusion}, we further demonstrate the benefit of our model by testing it on daylight videos where the activity suffers from local darkness caused by occlusion. For this we utilize the activity segmentation masks provided by \cite{darkhalil2022epic} for 182 video clips from the validation set of EPIC-Kitchens. We simulate occlusions by setting the pixel intensity of object regions to zero. While the visual-only model obtains 26.4\% accuracy, the vanilla audio-visual transformer~\citep{gabeur2020multi} results in 1.3\% improvement by the aid of sound. Our darkness-adaptive model gets a further 3.4\% increase since it adapts the audio-visual fusion based on how clear the activity is.

\subsection{Comparative Results}
Previous works recognizing activities in the dark~\citep{chen2021darklight,action02mcf,xu2020arid} require labeled dark videos for model learning, while our model does not. We instead compare our approach to test-time image enhancement, audio-visual fusion and domain adaptation works which are more similar to our task.

\noindent \textbf{Comparison with Image Enhancement}. 
We consider several state-of-the-art, low-light image enhancers \citep{zhang2019kindling,liu2021retinex,zhang2021beyond,ma2022toward,wang2018gladnet,wu2022uretinex,wang2022enhancement,zhou2022lednet,jin2022unsupervised}, as well as Gamma Intensity Correction~\citep{poynton2012digital}, a traditional method. Each is combined with a vanilla audio-visual transformer baseline~\citep{gabeur2020multi}. 
We show results for several image enhancers on EPIC-Kitchens in Table~\ref{tab:image_enhancement}. 
Image enhancement techniques degrade the performance of the vanilla transformer, due to the distortions they introduce (we present examples in the Appendix). 
In contrast, our model learns to generalize to dark videos without the need for image enhancement. 
Image enhancement techniques are also much slower than our model. 

\noindent \textbf{Comparison with Audio-Visual Fusion. }
We compare our model with recent state-of-the-art audio-visual fusion methods~\citep{lee2021crossattentional,tian2020unified,gabeur2020multi,nagrani2021attention} in Table~\ref{tab:alternative_fusion}. %
We use publicly available implementations by \cite{tian2020unified}, \cite{gabeur2020multi} and \cite{nagrani2021attention}, while re-implementing the method by \cite{lee2021crossattentional}. For all models, we use the same visual and audio encoders as ours. 
Although all existing methods are designed for the day, they still perform better than a visual-only model in the dark with up to +2.0\%, +2.7\% and -0.010 improvement. 
Nonetheless, our model better combines the complementary information from the two modalities in the dark, achieving +7.0\%, +5.2\% and -0.009 improvements over the best alternative. 
Although not our main goal, our model also gives the best audio-visual fusion for daylight videos. 
This is particularly noteworthy for Kinetics-Sound where it improves over the best baseline by +2.5\%.


\noindent \textbf{Comparison with Domain Adaptation.} 
In Table~\ref{tab:domain_adaptation}, we compare with three open-sourced unsupervised domain adaptation methods. While \cite{chen2019temporal} and \cite{munro2020multi} use the visual modality only, the state-of-the-art method by \cite{zhang2022audio} adopts both sight and sound. 
We apply these methods by treating the task-irrelevant, unlabeled dark videos as the target domain. 
For each we use the same visual and audio encoders as ours. As all of them are designed under the assumption that the unlabeled target domain data contain the same activities as the source, we outperform them by making effective use of task-irrelevant dark videos. 

\begin{table}[t]
\centering
\resizebox{1\linewidth}{!}{ 
\begin{tabular}{lcccccc}
\toprule
& \multicolumn{2}{c}{\textbf{EPIC-Kitchens}} \\
\cmidrule(lr){2-3} 
\textbf{Domain-Adaptation Model} & Dark $\uparrow$ & Day $\uparrow$ \\
\midrule
\rowcolor{visualblue}
\textbf{Visual only} & & \\
\cite{chen2019temporal} & 28.9 & 37.3 \\
\cite{munro2020multi} & 29.0 & 37.4 \\
\rowcolor{gray!20}
\textbf{Audio \& Visual} & & \\
\cite{zhang2022audio}  & 29.7 & 37.5  \\
\textbf{\textit{This paper}} & 35.6 & 37.6 \\
\bottomrule
\end{tabular}
}
\vspace{-0.7em}
\renewcommand\thetable{10}
\caption{\textbf{Comparison with Domain Adaptation}. While domain adaptation methods need unlabeled target domain data to contain the same activities as the source, ours effectively uses task-irrelevant videos to adapt to the dark. }
\label{tab:domain_adaptation}
\end{table}

\section{Conclusion} 
The day2dark gap exists in current activity recognizers because of the lack of dark training data and the low color contrast. 
Nevertheless, the lack of data can be compensated by our proposed supervision beyond daylight, which makes effective use of unlabeled task-irrelevant dark videos. 
The loss of information and distribution shift caused by the low color contrast in the dark can also be eliminated by our darkness-adaptive activity recognizer, which fuses sight with complementary information within sound adaptively to the illumination. 
Experiments show our proposals reduce the day2dark gap considerably and provide better generalization to dark environments than image enhancement, domain adaptation and audio-visual fusion approaches. %

\vspace{1em} 
\noindent \textbf{Data Availability Statement.} 
The datasets utilized during the current study are available in the internet repository: EPIC-Kitchens~\citep{EPIC}: \url{https://epic-kitchens.github.io/2023}; Kinetics-Sound~\citep{KineticsSound}: \url{https://www.deepmind.com/open-source/kinetics}; Charades~\citep{sigurdsson2016hollywood}: \url{https://prior.allenai.org/projects/charades}; YFCC100M~\citep{thomee2016yfcc100m}: \url{http://www.multimediacommons.org/}. We will provide the usage of these datasets on our project website: \url{https://xiaobai1217.github.io/Day2Dark}. 

\vspace{1em} 
\noindent \textbf{Acknowledgement.} 
This work is financially supported by the Inception Institute of Artificial Intelligence, the University of Amsterdam and the allowance Top consortia for Knowledge and Innovation (TKIs) from the Netherlands Ministry of Economic Affairs and Climate Policy.


\bibliographystyle{spbasic}      
\bibliography{main}   

\begin{thebibliography}{89}
\providecommand{\natexlab}[1]{#1}
\providecommand{\url}[1]{{#1}}
\providecommand{\urlprefix}{URL }
\expandafter\ifx\csname urlstyle\endcsname\relax
  \providecommand{\doi}[1]{DOI~\discretionary{}{}{}#1}\else
  \providecommand{\doi}{DOI~\discretionary{}{}{}\begingroup
  \urlstyle{rm}\Url}\fi
\providecommand{\eprint}[2][]{\url{#2}}

\bibitem[{Anderson et~al.(1996)Anderson, Motta, Chandrasekar, and
  Stokes}]{anderson1996proposal}
Anderson M, Motta R, Chandrasekar S, Stokes M (1996) Proposal for a standard
  default color space for the internet—srgb. In: Color and Imaging Conference

\bibitem[{Arandjelovi\'c and Zisserman(2017)}]{KineticsSound}
Arandjelovi\'c R, Zisserman A (2017) Look, listen and learn. In: ICCV

\bibitem[{Bain et~al.(2021)Bain, Nagrani, Varol, and
  Zisserman}]{bain2021frozen}
Bain M, Nagrani A, Varol G, Zisserman A (2021) Frozen in time: A joint video
  and image encoder for end-to-end retrieval. In: ICCV

\bibitem[{Carreira and Zisserman(2017)}]{Kinetics}
Carreira J, Zisserman A (2017) Quo vadis, action recognition? {A} new model and
  the kinetics dataset. In: CVPR

\bibitem[{Chen et~al.(2019{\natexlab{a}})Chen, Chen, Do, and
  Koltun}]{Chen_2019_ICCV}
Chen C, Chen Q, Do MN, Koltun V (2019{\natexlab{a}}) Seeing motion in the dark.
  In: ICCV

\bibitem[{Chen et~al.(2020)Chen, Xie, Vedaldi, and Zisserman}]{VGGSound}
Chen H, Xie W, Vedaldi A, Zisserman A (2020) {VGGSound}: a large-scale
  audio-visual dataset. In: ICASSP

\bibitem[{Chen et~al.(2019{\natexlab{b}})Chen, Kira, AlRegib, Yoo, Chen, and
  Zheng}]{chen2019temporal}
Chen MH, Kira Z, AlRegib G, Yoo J, Chen R, Zheng J (2019{\natexlab{b}})
  Temporal attentive alignment for large-scale video domain adaptation. In:
  ICCV

\bibitem[{Chen et~al.(2021)Chen, Chen, Liang, Gao, and Lin}]{chen2021darklight}
Chen R, Chen J, Liang Z, Gao H, Lin S (2021) Darklight networks for action
  recognition in the dark. In: CVPR Workshops

\bibitem[{Choi et~al.(2020)Choi, Sharma, Schulter, and Huang}]{choi2020shuffle}
Choi J, Sharma G, Schulter S, Huang JB (2020) Shuffle and attend: Video domain
  adaptation. In: ECCV

\bibitem[{Damen et~al.(2022)Damen, Doughty, Farinella, , Furnari, Ma, Kazakos,
  Moltisanti, Munro, Perrett, Price, and Wray}]{EPIC}
Damen D, Doughty H, Farinella GM, , Furnari A, Ma J, Kazakos E, Moltisanti D,
  Munro J, Perrett T, Price W, Wray M (2022) Rescaling egocentric vision:
  Collection, pipeline and challenges for epic-kitchens-100. IJCV 130:33–55,
  \urlprefix\url{https://doi.org/10.1007/s11263-021-01531-2}

\bibitem[{Darkhalil et~al.(2022)Darkhalil, Shan, Zhu, Ma, Kar, Higgins, Fidler,
  Fouhey, and Damen}]{darkhalil2022epic}
Darkhalil A, Shan D, Zhu B, Ma J, Kar A, Higgins REL, Fidler S, Fouhey D, Damen
  D (2022) Epic-kitchens visor benchmark: Video segmentations and object
  relations. In: NeurIPS Datasets and Benchmarks Track

\bibitem[{Dosovitskiy et~al.(2020)Dosovitskiy, Beyer, Kolesnikov, Weissenborn,
  Zhai, Unterthiner, Dehghani, Minderer, Heigold, Gelly et~al.}]{vit}
Dosovitskiy A, Beyer L, Kolesnikov A, Weissenborn D, Zhai X, Unterthiner T,
  Dehghani M, Minderer M, Heigold G, Gelly S, et~al. (2020) An image is worth
  16x16 words: Transformers for image recognition at scale. In: ICLR

\bibitem[{Doughty and Snoek(2022)}]{doughty2022you}
Doughty H, Snoek CGM (2022) How do you do it? fine-grained action understanding
  with pseudo-adverbs. In: CVPR

\bibitem[{Feichtenhofer et~al.(2019)Feichtenhofer, Fan, Malik, and
  He}]{slowfast}
Feichtenhofer C, Fan H, Malik J, He K (2019) {SlowFast} networks for video
  recognition. In: ICCV

\bibitem[{Gabeur et~al.(2020)Gabeur, Sun, Alahari, and
  Schmid}]{gabeur2020multi}
Gabeur V, Sun C, Alahari K, Schmid C (2020) Multi-modal transformer for video
  retrieval. In: ECCV

\bibitem[{Gan et~al.(2019)Gan, Zhao, Chen, Cox, and Torralba}]{gan2019self}
Gan C, Zhao H, Chen P, Cox D, Torralba A (2019) Self-supervised moving vehicle
  tracking with stereo sound. In: ICCV

\bibitem[{Gao et~al.(2016)Gao, Du, Liu, Lv, Yang, Meng, and
  Hauptmann}]{gao2016infar}
Gao C, Du Y, Liu J, Lv J, Yang L, Meng D, Hauptmann AG (2016) {InfAR} dataset:
  Infrared action recognition at different times. Neurocomputing 212:36--47

\bibitem[{Gao et~al.(2022)Gao, Guo, Wang, and Zhang}]{gao2022cross}
Gao H, Guo J, Wang G, Zhang Q (2022) Cross-domain correlation distillation for
  unsupervised domain adaptation in nighttime semantic segmentation. In: CVPR

\bibitem[{Gao et~al.(2020)Gao, Oh, Grauman, and Torresani}]{gao2020listen}
Gao R, Oh TH, Grauman K, Torresani L (2020) Listen to look: Action recognition
  by previewing audio. In: CVPR

\bibitem[{Gavrilyuk et~al.(2021)Gavrilyuk, Jain, Karmanov, and
  Snoek}]{gavrilyuk2021motionaugmented}
Gavrilyuk K, Jain M, Karmanov I, Snoek CGM (2021) Motion-augmented
  self-training for video recognition at smaller scale. In: ICCV

\bibitem[{Gebhardt and Wolf(2018)}]{gebhardt2018camel}
Gebhardt E, Wolf M (2018) {CAMEL} dataset for visual and thermal infrared
  multiple object detection and tracking. In: AVSS

\bibitem[{Girdhar et~al.(2022)Girdhar, Singh, Ravi, van~der Maaten, Joulin, and
  Misra}]{girdhar2022omnivore}
Girdhar R, Singh M, Ravi N, van~der Maaten L, Joulin A, Misra I (2022)
  Omnivore: A single model for many visual modalities. In: CVPR

\bibitem[{He et~al.(2016)He, Zhang, Ren, and Sun}]{resnet}
He K, Zhang X, Ren S, Sun J (2016) Deep residual learning for image
  recognition. In: CVPR

\bibitem[{Hu et~al.(2020)Hu, Mou, Wang, Gao, Hua, Dou, and Zhu}]{crowdcounting}
Hu D, Mou L, Wang Q, Gao J, Hua Y, Dou D, Zhu XX (2020) Ambient sound helps:
  Audiovisual crowd counting in extreme conditions. In: ECCV

\bibitem[{Jamal et~al.(2018)Jamal, Namboodiri, Deodhare, and
  Venkatesh}]{jamal2018deep}
Jamal A, Namboodiri VP, Deodhare D, Venkatesh K (2018) Deep domain adaptation
  in action space. In: BMVC

\bibitem[{Jia et~al.(2022)Jia, Tang, Chen, Cardie, Belongie, Hariharan, and
  Lim}]{jia2022visual}
Jia M, Tang L, Chen BC, Cardie C, Belongie S, Hariharan B, Lim SN (2022) Visual
  prompt tuning. ECCV

\bibitem[{Jiang and Zheng(2019)}]{jiang2019learning}
Jiang H, Zheng Y (2019) Learning to see moving objects in the dark. In: ICCV

\bibitem[{Jiang et~al.(2017)Jiang, Rozgic, and Adali}]{jiang2017learning}
Jiang Z, Rozgic V, Adali S (2017) Learning spatiotemporal features for infrared
  action recognition with 3d convolutional neural networks. In: CVPR Workshops

\bibitem[{Jin et~al.(2022)Jin, Yang, and Tan}]{jin2022unsupervised}
Jin Y, Yang W, Tan RT (2022) Unsupervised night image enhancement: When layer
  decomposition meets light-effects suppression. In: ECCV

\bibitem[{Kim et~al.(2021)Kim, Tsai, Zhuang, Yu, Sclaroff, Saenko, and
  Chandraker}]{iccv2021videoadaptation}
Kim D, Tsai YH, Zhuang B, Yu X, Sclaroff S, Saenko K, Chandraker M (2021)
  Learning cross-modal contrastive features for video domain adaptation. In:
  ICCV

\bibitem[{Korbar et~al.(2018)Korbar, Tran, and
  Torresani}]{korbar2018cooperative}
Korbar B, Tran D, Torresani L (2018) Cooperative learning of audio and video
  models from self-supervised synchronization. In: NeurIPS

\bibitem[{Korbar et~al.(2019)Korbar, Tran, and Torresani}]{korbar2019scsampler}
Korbar B, Tran D, Torresani L (2019) {SCSampler}: Sampling salient clips from
  video for efficient action recognition. In: ICCV

\bibitem[{Kuehne et~al.(2011)Kuehne, Jhuang, Garrote, Poggio, and
  Serre}]{kuehne2011hmdb}
Kuehne H, Jhuang H, Garrote E, Poggio T, Serre T (2011) Hmdb: a large video
  database for human motion recognition. In: ICCV

\bibitem[{Lee et~al.(2013)}]{lee2013pseudo}
Lee DH, et~al. (2013) Pseudo-label: The simple and efficient semi-supervised
  learning method for deep neural networks. In: ICLR Workshops

\bibitem[{Lee et~al.(2021)Lee, Jain, Park, and Yun}]{lee2021crossattentional}
Lee JT, Jain M, Park H, Yun S (2021) Cross-attentional audio-visual fusion for
  weakly-supervised action localization. In: ICLR

\bibitem[{Li et~al.(2023)Li, Wang, and Cui}]{li2023decoupled}
Li Y, Wang Y, Cui Z (2023) Decoupled multimodal distilling for emotion
  recognition. In: CVPR

\bibitem[{Lin et~al.(2023)Lin, Sung, Lei, Bansal, and
  Bertasius}]{lin2023vision}
Lin YB, Sung YL, Lei J, Bansal M, Bertasius G (2023) Vision transformers are
  parameter-efficient audio-visual learners. In: CVPR

\bibitem[{Liu et~al.(2021{\natexlab{a}})Liu, Ma, Zhang, Fan, and
  Luo}]{liu2021retinex}
Liu R, Ma L, Zhang J, Fan X, Luo Z (2021{\natexlab{a}}) Retinex-inspired
  unrolling with cooperative prior architecture search for low-light image
  enhancement. In: CVPR

\bibitem[{Liu et~al.(2018)Liu, Lu, Li, Yang, and Yao}]{liu2018global}
Liu Y, Lu Z, Li J, Yang T, Yao C (2018) Global temporal representation based
  cnns for infrared action recognition. IEEE Signal Processing Letters
  25(6):848--852

\bibitem[{Liu et~al.(2021{\natexlab{b}})Liu, Lin, Cao, Hu, Wei, Zhang, Lin, and
  Guo}]{liu2021swin}
Liu Z, Lin Y, Cao Y, Hu H, Wei Y, Zhang Z, Lin S, Guo B (2021{\natexlab{b}})
  Swin transformer: Hierarchical vision transformer using shifted windows. In:
  ICCV

\bibitem[{Luo et~al.(2023)Luo, Wang, Yang, and Liu}]{luo2023similarity}
Luo R, Wang W, Yang W, Liu J (2023) Similarity min-max: Zero-shot day-night
  domain adaptation. In: ICCV

\bibitem[{Ma et~al.(2022)Ma, Ma, Liu, Fan, and Luo}]{ma2022toward}
Ma L, Ma T, Liu R, Fan X, Luo Z (2022) Toward fast, flexible, and robust
  low-light image enhancement. In: CVPR

\bibitem[{Marsza{\l}ek et~al.(2009)Marsza{\l}ek, Laptev, and
  Schmid}]{marszalek09}
Marsza{\l}ek M, Laptev I, Schmid C (2009) Actions in context. In: CVPR

\bibitem[{Miech et~al.(2019)Miech, Zhukov, Alayrac, Tapaswi, Laptev, and
  Sivic}]{miech2019howto100m}
Miech A, Zhukov D, Alayrac JB, Tapaswi M, Laptev I, Sivic J (2019) Howto100m:
  Learning a text-video embedding by watching hundred million narrated video
  clips. In: CVPR

\bibitem[{Mo and Morgado(2022)}]{mo2022localizing}
Mo S, Morgado P (2022) Localizing visual sounds the easy way. In: ECCV

\bibitem[{Munro and Damen(2020)}]{munro2020multi}
Munro J, Damen D (2020) Multi-modal domain adaptation for fine-grained action
  recognition. In: CVPR

\bibitem[{Nagrani et~al.(2021)Nagrani, Yang, Arnab, Jansen, Schmid, and
  Sun}]{nagrani2021attention}
Nagrani A, Yang S, Arnab A, Jansen A, Schmid C, Sun C (2021) Attention
  bottlenecks for multimodal fusion. In: NeurIPS

\bibitem[{Neumann et~al.(2018)Neumann, Karg, Zhang, Scharfenberger, Piegert,
  Mistr, Prokofyeva, Thiel, Vedaldi, Zisserman, and
  Schiele}]{neumann2018nightowls}
Neumann L, Karg M, Zhang S, Scharfenberger C, Piegert E, Mistr S, Prokofyeva O,
  Thiel R, Vedaldi A, Zisserman A, Schiele B (2018) Nightowls: A pedestrians at
  night dataset. In: ACCV

\bibitem[{Pan et~al.(2020)Pan, Cao, Adeli, and Niebles}]{pan2020adversarial}
Pan B, Cao Z, Adeli E, Niebles JC (2020) Adversarial cross-domain action
  recognition with co-attention. In: AAAI

\bibitem[{Poynton(2012)}]{poynton2012digital}
Poynton C (2012) Digital video and HD: Algorithms and Interfaces. Elsevier

\bibitem[{Rahman et~al.(2019)Rahman, Xu, and Sigal}]{rahman2019watch}
Rahman T, Xu B, Sigal L (2019) Watch, listen and tell: Multi-modal weakly
  supervised dense event captioning. In: ICCV

\bibitem[{Sakaridis et~al.(2019)Sakaridis, Dai, and Gool}]{sakaridis2019guided}
Sakaridis C, Dai D, Gool LV (2019) Guided curriculum model adaptation and
  uncertainty-aware evaluation for semantic nighttime image segmentation. In:
  CVPR

\bibitem[{Shvetsova et~al.(2022)Shvetsova, Chen, Rouditchenko, Thomas,
  Kingsbury, Feris, Harwath, Glass, and Kuehne}]{shvetsova2022everything}
Shvetsova N, Chen B, Rouditchenko A, Thomas S, Kingsbury B, Feris RS, Harwath
  D, Glass J, Kuehne H (2022) Everything at once-multi-modal fusion transformer
  for video retrieval. In: CVPR

\bibitem[{Sigurdsson et~al.(2016)Sigurdsson, Varol, Wang, Farhadi, Laptev, and
  Gupta}]{sigurdsson2016hollywood}
Sigurdsson GA, Varol G, Wang X, Farhadi A, Laptev I, Gupta A (2016) Hollywood
  in homes: Crowdsourcing data collection for activity understanding. In: ECCV

\bibitem[{Sigurdsson et~al.(2018)Sigurdsson, Gupta, Schmid, Farhadi, and
  Alahari}]{sigurdsson2018actor}
Sigurdsson GA, Gupta A, Schmid C, Farhadi A, Alahari K (2018) Actor and
  observer: Joint modeling of first and third-person videos. In: CVPR

\bibitem[{Song et~al.(2021)Song, Zhao, Yang, Yue, Xu, Hu, and
  Chai}]{song2021spatio}
Song X, Zhao S, Yang J, Yue H, Xu P, Hu R, Chai H (2021) Spatio-temporal
  contrastive domain adaptation for action recognition. In: CVPR

\bibitem[{Sun et~al.(2019)Sun, Wang, Yang, and Xiang}]{sun2019see}
Sun L, Wang K, Yang K, Xiang K (2019) See clearer at night: towards robust
  nighttime semantic segmentation through day-night image conversion. In:
  Artificial Intelligence and Machine Learning in Defense Applications, vol
  11169, pp 77--89

\bibitem[{Thomee et~al.(2016)Thomee, Shamma, Friedland, Elizalde, Ni, Poland,
  Borth, and Li}]{thomee2016yfcc100m}
Thomee B, Shamma DA, Friedland G, Elizalde B, Ni K, Poland D, Borth D, Li LJ
  (2016) Yfcc100m: The new data in multimedia research. Communications of the
  ACM 59(2):64--73

\bibitem[{Tian et~al.(2018)Tian, Shi, Li, Duan, and Xu}]{tian2018audio}
Tian Y, Shi J, Li B, Duan Z, Xu C (2018) Audio-visual event localization in
  unconstrained videos. In: ECCV

\bibitem[{Tian et~al.(2019)Tian, Guan, Goodman, Moore, and
  Xu}]{tian2018attempt}
Tian Y, Guan C, Goodman J, Moore M, Xu C (2019) An attempt towards
  interpretable audio-visual video captioning. In: ICCV

\bibitem[{Tian et~al.(2020)Tian, Li, and Xu}]{tian2020unified}
Tian Y, Li D, Xu C (2020) Unified multisensory perception: weakly-supervised
  audio-visual video parsing. In: ECCV

\bibitem[{Ulhaq et~al.(2016)Ulhaq, Yin, Zhang, and Gondal}]{action02mcf}
Ulhaq A, Yin X, Zhang Y, Gondal I (2016) Action-02mcf: A robust space-time
  correlation filter for action recognition in clutter and adverse lighting
  conditions. In: ACIVS

\bibitem[{Valverde et~al.(2021)Valverde, Hurtado, and
  Valada}]{valverde2021there}
Valverde FR, Hurtado JV, Valada A (2021) There is more than meets the eye:
  Self-supervised multi-object detection and tracking with sound by distilling
  multimodal knowledge. In: CVPR

\bibitem[{Wang et~al.(2022{\natexlab{a}})Wang, Xu, and
  Lau}]{wang2022enhancement}
Wang H, Xu K, Lau RW (2022{\natexlab{a}}) Local color distributions prior for
  image enhancement. In: ECCV

\bibitem[{Wang et~al.(2022{\natexlab{b}})Wang, Lan, Liu, Ouyang, Qin, Lu, Chen,
  Zeng, and Yu}]{wang2022generalizing}
Wang J, Lan C, Liu C, Ouyang Y, Qin T, Lu W, Chen Y, Zeng W, Yu P
  (2022{\natexlab{b}}) Generalizing to unseen domains: A survey on domain
  generalization. IEEE Transactions on Knowledge and Data Engineering

\bibitem[{Wang and Deng(2018)}]{wang2018deep}
Wang M, Deng W (2018) Deep visual domain adaptation: A survey. Neurocomputing
  312:135--153

\bibitem[{Wang et~al.(2018{\natexlab{a}})Wang, Wei, Yang, and
  Liu}]{wang2018gladnet}
Wang W, Wei C, Yang W, Liu J (2018{\natexlab{a}}) Gladnet: Low-light
  enhancement network with global awareness. In: 13th IEEE International
  Conference on Automatic Face \& Gesture Recognition

\bibitem[{Wang et~al.(2018{\natexlab{b}})Wang, Wang, and Wang}]{wang2018watch}
Wang X, Wang YF, Wang WY (2018{\natexlab{b}}) Watch, listen, and describe:
  Globally and locally aligned cross-modal attentions for video captioning. In:
  NAACL-HLT

\bibitem[{Wang et~al.(2022{\natexlab{c}})Wang, Zhang, Lee, Zhang, Sun, Ren, Su,
  Perot, Dy, and Pfister}]{promptcontinual}
Wang Z, Zhang Z, Lee CY, Zhang H, Sun R, Ren X, Su G, Perot V, Dy J, Pfister T
  (2022{\natexlab{c}}) Learning to prompt for continual learning. In: CVPR

\bibitem[{Wei et~al.(2018)Wei, Wang, Yang, and Liu}]{wei2018deep}
Wei C, Wang W, Yang W, Liu J (2018) Deep retinex decomposition for low-light
  enhancement. In: BMVC

\bibitem[{Wu et~al.(2022{\natexlab{a}})Wu, Li, Mangalam, Fan, Xiong, Malik, and
  Feichtenhofer}]{wu2022memvit}
Wu CY, Li Y, Mangalam K, Fan H, Xiong B, Malik J, Feichtenhofer C
  (2022{\natexlab{a}}) Memvit: Memory-augmented multiscale vision transformer
  for efficient long-term video recognition. In: CVPR

\bibitem[{Wu et~al.(2020)Wu, Liu, Shi, Sun, Shao, Wu, and Yang}]{xdviolence}
Wu P, Liu J, Shi Y, Sun Y, Shao F, Wu Z, Yang Z (2020) Not only look, but also
  listen: Learning multimodal violence detection under weak supervision. In:
  ECCV

\bibitem[{Wu et~al.(2022{\natexlab{b}})Wu, Weng, Zhang, Wang, Yang, and
  Jiang}]{wu2022uretinex}
Wu W, Weng J, Zhang P, Wang X, Yang W, Jiang J (2022{\natexlab{b}})
  Uretinex-net: Retinex-based deep unfolding network for low-light image
  enhancement. In: CVPR

\bibitem[{Wu and Yang(2021)}]{wu2021exploring}
Wu Y, Yang Y (2021) Exploring heterogeneous clues for weakly-supervised
  audio-visual video parsing. In: CVPR

\bibitem[{Wu et~al.(2019)Wu, Zhu, Yan, and Yang}]{wu2019dual}
Wu Y, Zhu L, Yan Y, Yang Y (2019) Dual attention matching for audio-visual
  event localization. In: ICCV

\bibitem[{Xu et~al.(2021{\natexlab{a}})Xu, Yang, Cao, Chen, Li, and
  Mao}]{xu2021partial}
Xu Y, Yang J, Cao H, Chen Z, Li Q, Mao K (2021{\natexlab{a}}) Partial video
  domain adaptation with partial adversarial temporal attentive network. In:
  ICCV

\bibitem[{Xu et~al.(2021{\natexlab{b}})Xu, Yang, Cao, Mao, Yin, and
  See}]{xu2020arid}
Xu Y, Yang J, Cao H, Mao K, Yin J, See S (2021{\natexlab{b}}) Arid: A new
  dataset for recognizing action in the dark. International Workshop on Deep
  Learning for Human Activity Recognition

\bibitem[{Xu et~al.(2021{\natexlab{c}})Xu, Yang, Cao, Wu, Wu, Zhao, and
  Chen}]{xu2021multi}
Xu Y, Yang J, Cao H, Wu K, Wu M, Zhao R, Chen Z (2021{\natexlab{c}})
  Multi-source video domain adaptation with temporal attentive moment
  alignment. arXiv preprint arXiv:210909964

\bibitem[{Xu et~al.(2022)Xu, Yang, Cao, Wu, Min, and Chen}]{xu2022learning}
Xu Y, Yang J, Cao H, Wu K, Min W, Chen Z (2022) Source-free video domain
  adaptation by learning temporal consistency for action recognition. ECCV

\bibitem[{Yan et~al.(2022)Yan, Xiong, Arnab, Lu, Zhang, Sun, and
  Schmid}]{yan2022multiview}
Yan S, Xiong X, Arnab A, Lu Z, Zhang M, Sun C, Schmid C (2022) Multiview
  transformers for video recognition. In: CVPR

\bibitem[{Yang et~al.(2022)Yang, Huang, Sugano, and Sato}]{yang2022interact}
Yang L, Huang Y, Sugano Y, Sato Y (2022) Interact before align: Leveraging
  cross-modal knowledge for domain adaptive action recognition. In: CVPR

\bibitem[{Ye et~al.(2022)Ye, Fu, Zheng, Paudel, and Chen}]{ye2022unsupervised}
Ye J, Fu C, Zheng G, Paudel DP, Chen G (2022) Unsupervised domain adaptation
  for nighttime aerial tracking. In: CVPR

\bibitem[{Zhang et~al.(2018)Zhang, Cisse, Dauphin, and
  Lopez-Paz}]{zhang2017mixup}
Zhang H, Cisse M, Dauphin YN, Lopez-Paz D (2018) Mixup: Beyond empirical risk
  minimization. In: ICLR

\bibitem[{Zhang et~al.(2019)Zhang, Zhang, and Guo}]{zhang2019kindling}
Zhang Y, Zhang J, Guo X (2019) Kindling the darkness: A practical low-light
  image enhancer. In: ACMMM

\bibitem[{Zhang et~al.(2021{\natexlab{a}})Zhang, Guo, Ma, Liu, and
  Zhang}]{zhang2021beyond}
Zhang Y, Guo X, Ma J, Liu W, Zhang J (2021{\natexlab{a}}) Beyond brightening
  low-light images. IJCV 129(4):1013--1037

\bibitem[{Zhang et~al.(2021{\natexlab{b}})Zhang, Shao, and
  Snoek}]{zhang2021repetitive}
Zhang Y, Shao L, Snoek CGM (2021{\natexlab{b}}) Repetitive activity counting by
  sight and sound. In: CVPR

\bibitem[{Zhang et~al.(2022)Zhang, Doughty, Shao, and Snoek}]{zhang2022audio}
Zhang Y, Doughty H, Shao L, Snoek CGM (2022) Audio-adaptive activity
  recognition across video domains. In: CVPR

\bibitem[{Zhou et~al.(2022{\natexlab{a}})Zhou, Yang, Loy, and
  Liu}]{zhou2022learning}
Zhou K, Yang J, Loy CC, Liu Z (2022{\natexlab{a}}) Learning to prompt for
  vision-language models. IJCV pp 1--12

\bibitem[{Zhou et~al.(2022{\natexlab{b}})Zhou, Li, and Loy}]{zhou2022lednet}
Zhou S, Li C, Loy CC (2022{\natexlab{b}}) Lednet: Joint low-light enhancement
  and deblurring in the dark. In: ECCV

\end{thebibliography}

\clearpage
\appendix

\begin{table*}[t!]
\centering
\resizebox{0.73\linewidth}{!}{
    \begin{tabular}{lccccccc}
    \toprule
    & \multicolumn{2}{c}{\textbf{EPIC-Kitchens}} & \multicolumn{2}{c}{\textbf{Kinetics-Sound}} & \multicolumn{2}{c}{\textbf{Charades}}\\
    \cmidrule(lr){2-3} \cmidrule(lr){4-5} \cmidrule(lr){6-7}
\textbf{Model} & Dark $\uparrow$ & Day $\uparrow$ & Dark $\uparrow$ & Day $\uparrow$ & Dark $\downarrow$ & Day $\downarrow$ \\
\midrule
\rowcolor{audiogreen}
\textbf{Audio only} & & & & & &\\
Audio encoder & 12.8 & 10.8 & 51.5 & 59.3 & 0.585 & 0.554 \\
\rowcolor{visualblue}
\textbf{Visual only} & & & & & &\\
Visual encoder & 18.5 & 33.6 & 63.7 & 75.3 & 0.061 & 0.040 \\
+ Pseudo-supervised day2dark learning & 19.6 & 33.9 & 65.2 & 76.0 & 0.055 & 0.038 \\
+ Day2dark-mix finetuning & 20.1 & 34.0 & 66.1 & 76.4 & 0.052 & 0.037 \\
\rowcolor{gray!20}
\textbf{Audio \& Visual} & & & & & &\\
Darkness-adaptive activity recognizer & 22.2 & 35.7 & 68.1 & 79.8 & 0.048 & 0.036 \\
+ Pseudo-supervised day2dark learning & 27.1 & 37.2 & 74.4 & 82.9 & 0.041 & 0.033 \\
+ Day2dark-mix finetuning & 28.2 & 37.5 & 75.7 & 83.9 & 0.037 & 0.031 \\
\bottomrule
\end{tabular}
}
\vspace{-0.5em}
\caption{\textbf{Ablation of Supervision Beyond Daylight with a SlowFast backbone}. 
Pseudo-supervised day2dark learning and day2dark-mix improve accuracy of a visual SlowFast encoder in the dark and also improve performance in the day. They have an even bigger impact when training our darkness-adaptive recognizer. 
}
\vspace{-0.5em}
\label{tab:supervision_beyond_daylight_ablation_slowfast}
\end{table*}

\begin{table*}[h]
\centering
\resizebox{0.73\linewidth}{!}{
    \begin{tabular}{lcccccc}
    \toprule
    & \multicolumn{2}{c}{\textbf{EPIC-Kitchens}} & \multicolumn{2}{c}{\textbf{Kinetics-Sound}} & \multicolumn{2}{c}{\textbf{Charades}} \\
    \cmidrule(lr){2-3} \cmidrule(lr){4-5} \cmidrule(lr){6-7}
\textbf{Model} & Dark $\uparrow$ & Day $\uparrow$ & Dark $\uparrow$ & Day $\uparrow$ & Dark $\downarrow$ & Day $\downarrow$ \\
\midrule
\rowcolor{gray!20}
\textbf{Audio \& Visual} & & & & & & \\
Vanilla multi-modal transformer & 22.7 & 34.9 & 68.2 & 80.1 & 0.049 & 0.036 \\
+ Darkness probe \& adaptive encoder & 24.3 & 35.2 & 70.1 & 81.3 & 0.045 & 0.035 \\
+ Adaptive prompts & 27.1 & 36.8 & 74.3 & 82.9 & 0.040 & 0.032 \\
+ Adaptive classification  & 28.2 & 37.5 & 75.7 & 83.9 & 0.037 & 0.031 \\
\bottomrule
\end{tabular}
}
\vspace{-0.5em}
\caption{\textbf{Ablation of Darkness-Adaptive Activity Recognizer with a SlowFast backbone}. All model variants in this table are trained with supervision beyond daylight. 
Each component increases performance in the dark and the day. 
}
\label{tab:darkness_fusion_ablation_slowfast}
\vspace{-0.8em}
\end{table*}

\section{Activities Benefiting from Our Model}

\begin{figure}[t!]
\centering 
\includegraphics[width=1\linewidth,height=0.7\linewidth]{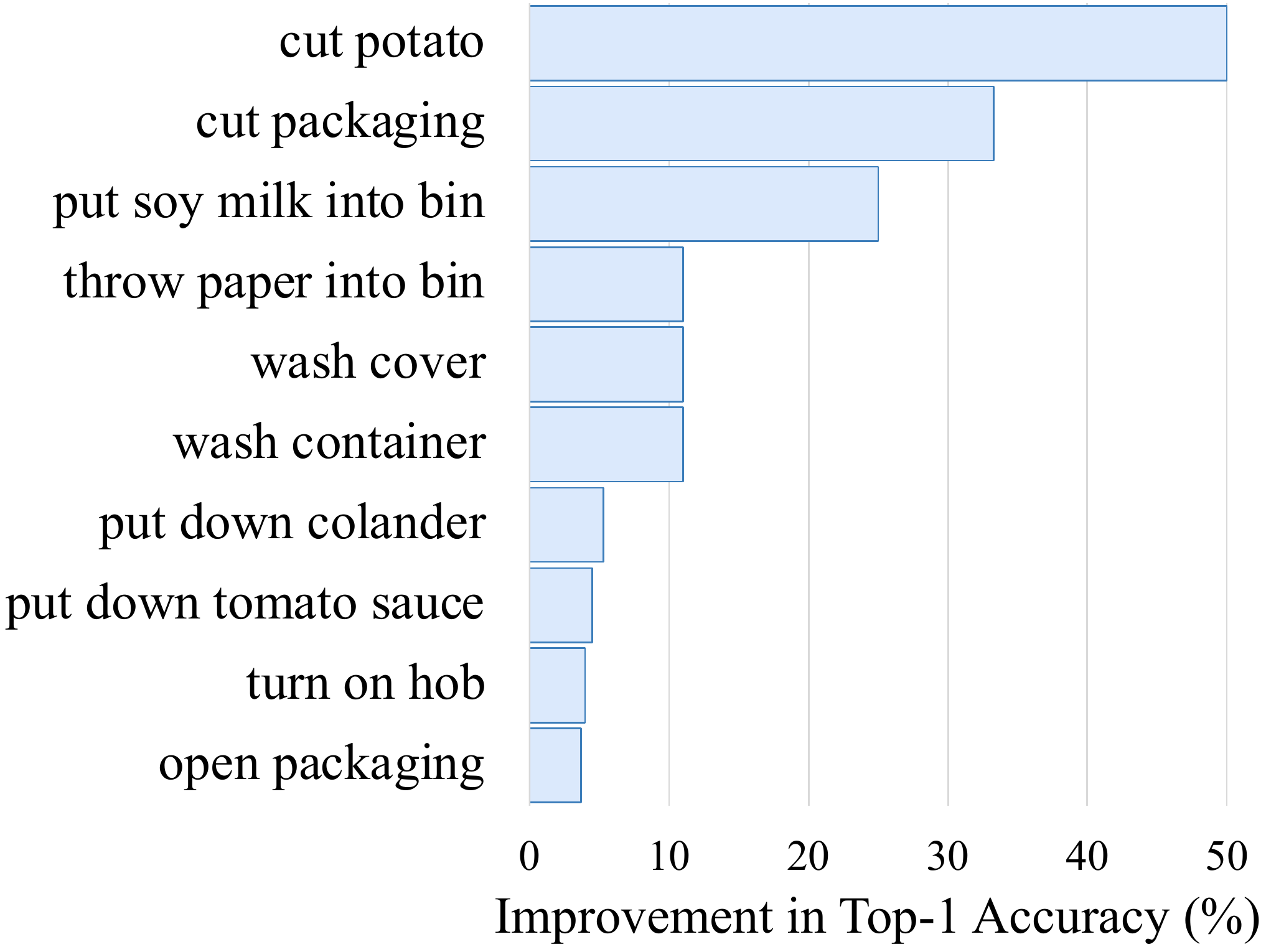}
\vspace{-1.5em}
\caption{\textbf{Activities in EPIC-Kitchens Which Benefit from Our Model} with Swin-T backbone. 
Sound is helpful in the dark even though not all activities have characteristic sound signals. 
}
\label{fig:acc_diff_epic}
\end{figure}

\begin{figure}[t!]
\centering 
\includegraphics[width=1\linewidth,height=0.65\linewidth]{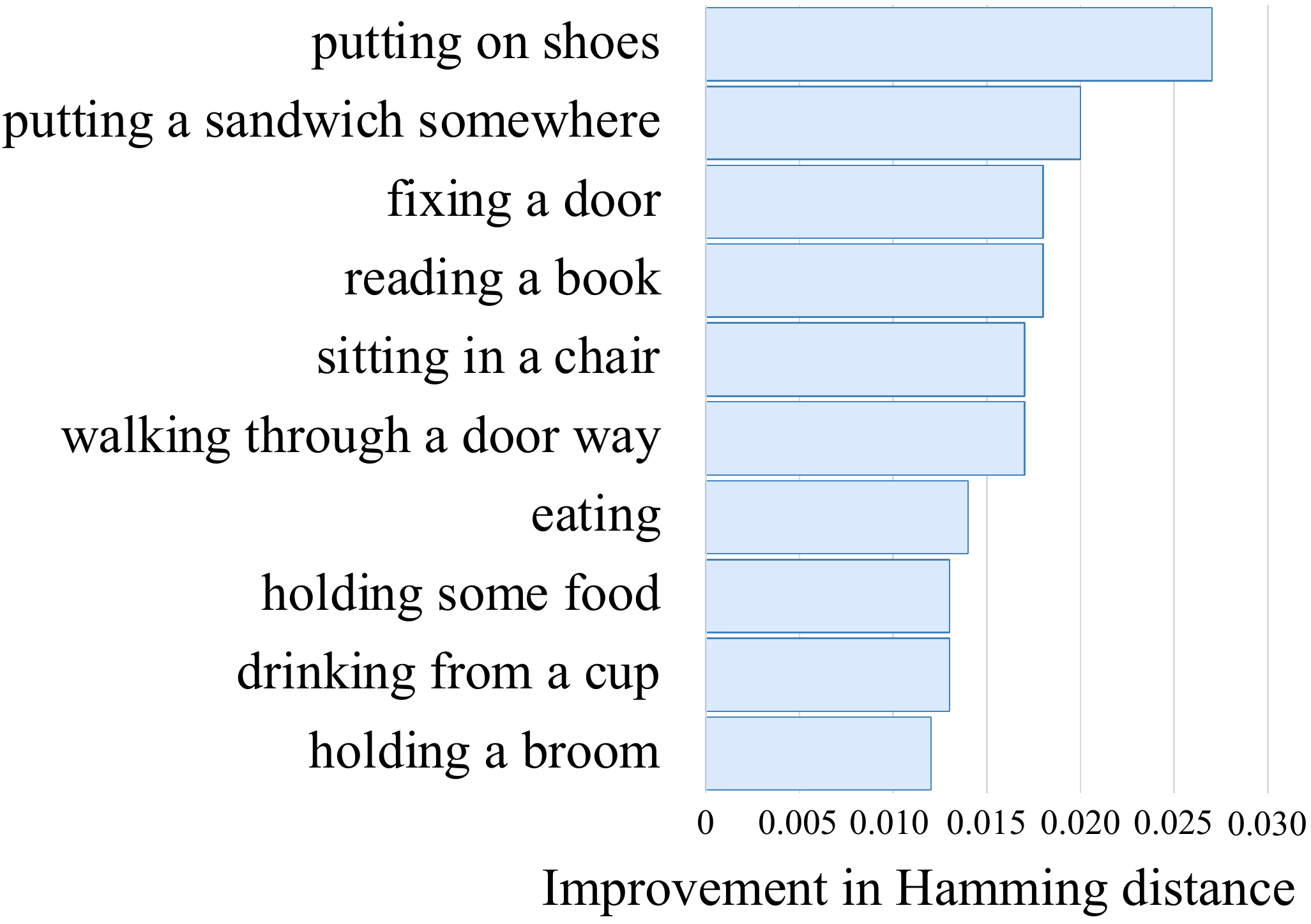}
\vspace{-1.5em}
\caption{\textbf{Activities in Charades Which Benefit from Our Model} with Swin-T backbone. 
We report the improvement in terms of hamming distance. 
Sound is helpful in the dark even though not all activities have characteristic sound signals. 
}
\label{fig:acc_diff2}
\end{figure}

In Figures~\ref{fig:acc_diff_epic} and~\ref{fig:acc_diff2}, we present activities that benefit most from our darkness-adaptive components over a visual-only baseline on EPIC-Kitchens and Charades~\citep{sigurdsson2016hollywood}. 
We observe that our model improves over both audible (\eg `fixing a door', and `eating') and silent (\eg `holding a broom' and `sitting in a chair') activities. 
This is because for silent activities, the environmental sound can also provide useful activity information. 
For example, when the environment is silent, the activity is impossible to be audible activities like `cut potato'. 
Overall, we conclude that audio is helpful for handling low-light conditions even for activities that do not have a characteristic sound. 

\section{Results with a SlowFast Backbone}

In the main paper, we demonstrate how our model reduces the day2dark gap when using a transformer backbone (Swin-T). 
Since the day2dark gap is also a problem for other architectures, we also apply our darkness-adaptive designs on a state-of-the-art convolutional backbone SlowFast~\citep{slowfast}. 
Ablations of our supervision beyond daylight and darkness-adaptive activity recognizer with SlowFast are shown in Table~\ref{tab:supervision_beyond_daylight_ablation_slowfast} and Table~\ref{tab:darkness_fusion_ablation_slowfast}. 
Specifically, our full model outperforms the visual-only encoder by +9.7\%, +9.6\% and -0.024 (hamming distance) in the dark on EPIC-Kitchens, Kinetics-Sound and Charades. Similar to the results of Swin-T in the main paper, each component contributes to the performance improvement. With the SlowFast backbone, our model also improves recognition in the day by a large amount, with +3.5\%, +7.5\% and -0.009 respective improvements on EPIC-Kitchens, Kinetics-Sound and Charades. 
%

\section{Ablation of Supervision Beyond Daylight}

\noindent \textbf{Two-Stage Training \textit{vs.} End-to-End. }
In our proposed supervision beyond daylight, we train the activity recognizer at two stages, \ie  pseudo-supervised day2dark learning followed by day2dark-mix finetuning. 
Alternatively, we can train the model end-to-end with the pseudo-supervised learning and day2dark-mix simultaneously. 
Then, the learning objective in Eq.~\ref{eq:pseudo_supervised} becomes:
\begin{equation}
    L = L_{\text{CE}} + \lambda L_{\mathcal{U}} + L_{\text{mix}}, 
\end{equation}
where $L_{\text{mix}}$ indicates the cross-entropy classification loss computed on the day2dark-mix samples. 
On EPIC-Kitchens, we obtain 35.4\% (dark) and 34.9\% (daylight), compared to 35.6\% and 37.6\% by two-stage training. 
Training end-to-end gives us a similar performance for dark videos, but hurts the performance on daylight videos. 
This is because combining the two stages lets the model focus too much on dark videos during training. 

\section{Ablation of Darkness-Adaptive Recognizer}

\begin{figure*}[t!]
\centering 
\includegraphics[width=\linewidth,height=1.1\linewidth]{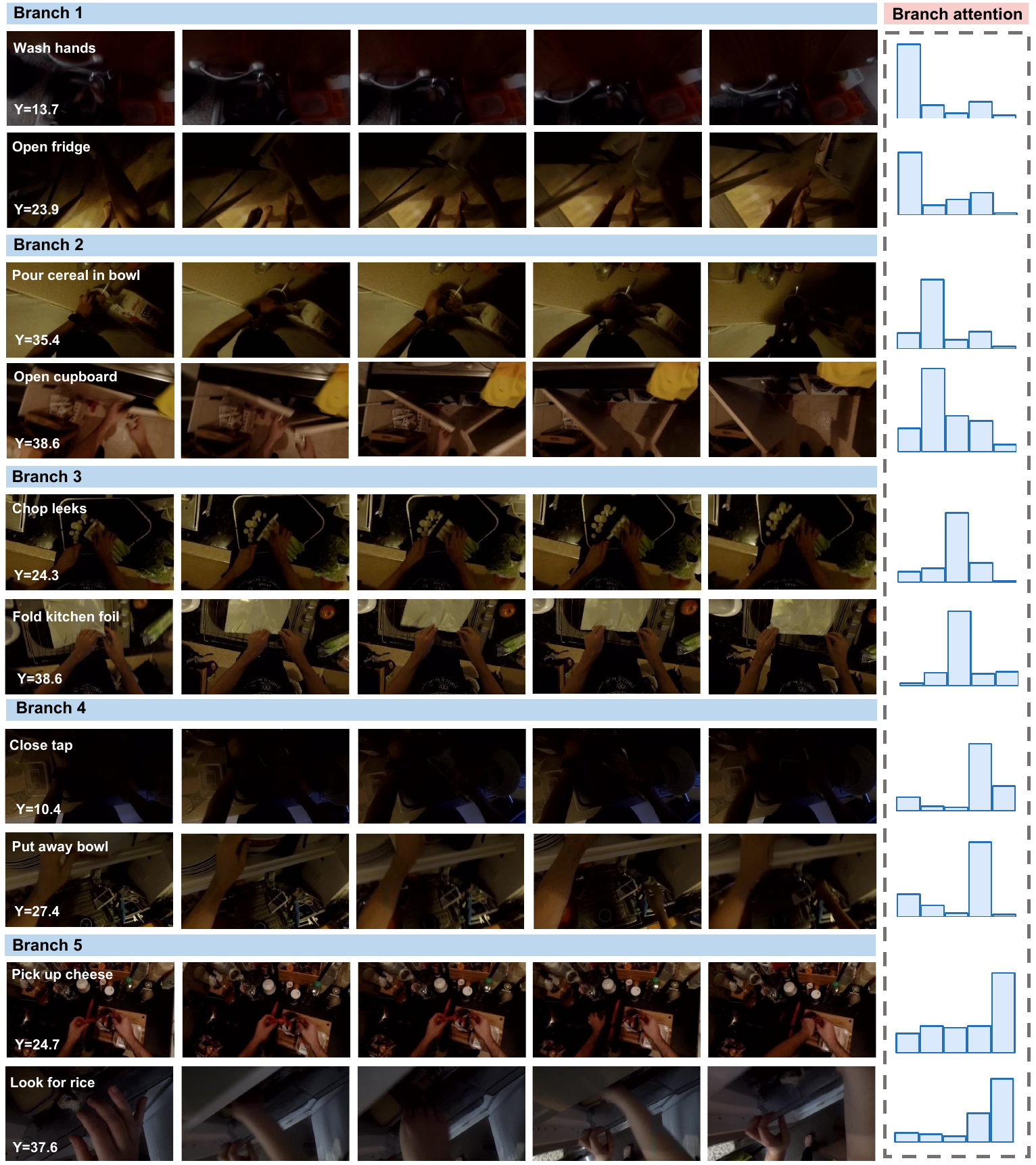}
\vspace{-1.2em}
\caption{\textbf{Examples of Branch Attention}. For each of the $K=5$ branches in our darkness-adaptive activity recognizer, we show two videos with high attention. We observe that each branch prefers a certain type of dark video. For instance, branch one attends when motion and sound are more reliable than the visual appearance. In contrast, branch three prefers activities that have clearer appearance and sound but little motion. We also show these examples in the supplementary video. }
\label{fig:branch_att}
\end{figure*}

\begin{table*}[t!]
\centering
\setlength{\tabcolsep}{3.5pt}
\resizebox{0.6\linewidth}{!} {
\begin{tabular}{lcccccc}
\toprule
\textbf{Percentage task-relevant} & \multicolumn{2}{c}{\textbf{0\%}} & \multicolumn{2}{c}{\textbf{10\%}} & \multicolumn{2}{c}{\textbf{20\%}} \\
\midrule
& Dark & Day & Dark & Day & Dark & Day \\
\cmidrule(lr){2-3} \cmidrule(lr){4-5} \cmidrule(lr){6-7}
Darkness-adaptive recognizer & 80.3 & 85.7 & 80.3 & 85.7 & 80.3 & 85.7\\
+ Pseudo-supervised day2dark & 84.2 & 86.7 & 84.3 & 86.5 & 84.2 & 86.8 \\
+ Day2dark-mix & 85.1 & 87.2 & 85.2 & 87.0 & 85.2 & 87.1 \\
\bottomrule
\end{tabular}
}
\vspace{-0.8em}
\caption{\textbf{Ablation of Percentage of Task-Relevant Dark Videos} on Kinetics-Sound. Regardless of the percentage of unlabeled task-relevant dark videos used, our proposed supervision beyond daylight is more effective than using our darkness-adaptive activity recognizer alone. }
\label{tab:appendix_percentage}
\end{table*}

\noindent \textbf{Percentage of task-relevant dark videos.} 
Our model is not reliant on all unlabeled dark videos being task-irrelevant. 
We set this constraint in the main paper to ensure our model works in this extreme since task-relevant dark videos can be rare. 
In Table~\ref{tab:appendix_percentage}, we examine when 10\% and 20\% of unlabeled videos may be relevant to Kinetics-Sound. 
Regardless of the percentage of task-relevant videos used, our proposed supervision is more effective than using our darkness-adaptive activity recognizer alone. 
The performance of our model is also insensitive to the percentage of task-relevant videos. 
This is because our model does not use these dark videos to directly learn to recognize the activities for the target task. 
Instead, we use these dark videos to provide the data distribution in low light conditions with pseudo labels of the auxiliary tasks for model learning. 
Therefore, the percentage of unlabeled task-relevant dark videos does not influence our model performance much. 
On EPIC-Kitchens, with 2.1\% task-relevant videos for training (the maximum we can obtain), the pseudo-supervised learning improves our recognizer from 29.3\%/37.4\% dark/day accuracy to 34.2\%/37.6\%. 
Adding the day2dark-mix improves the performance further to 35.6\%/37.7\%. This is comparable to using only task-irrelevant videos (37.7\%/37.6\%). 
Thus, our proposed supervision beyond daylight works regardless of whether task-relevant videos are included.

\begin{figure*}[t!]
\centering 
\includegraphics[width=0.9\linewidth,height=1.05\linewidth]{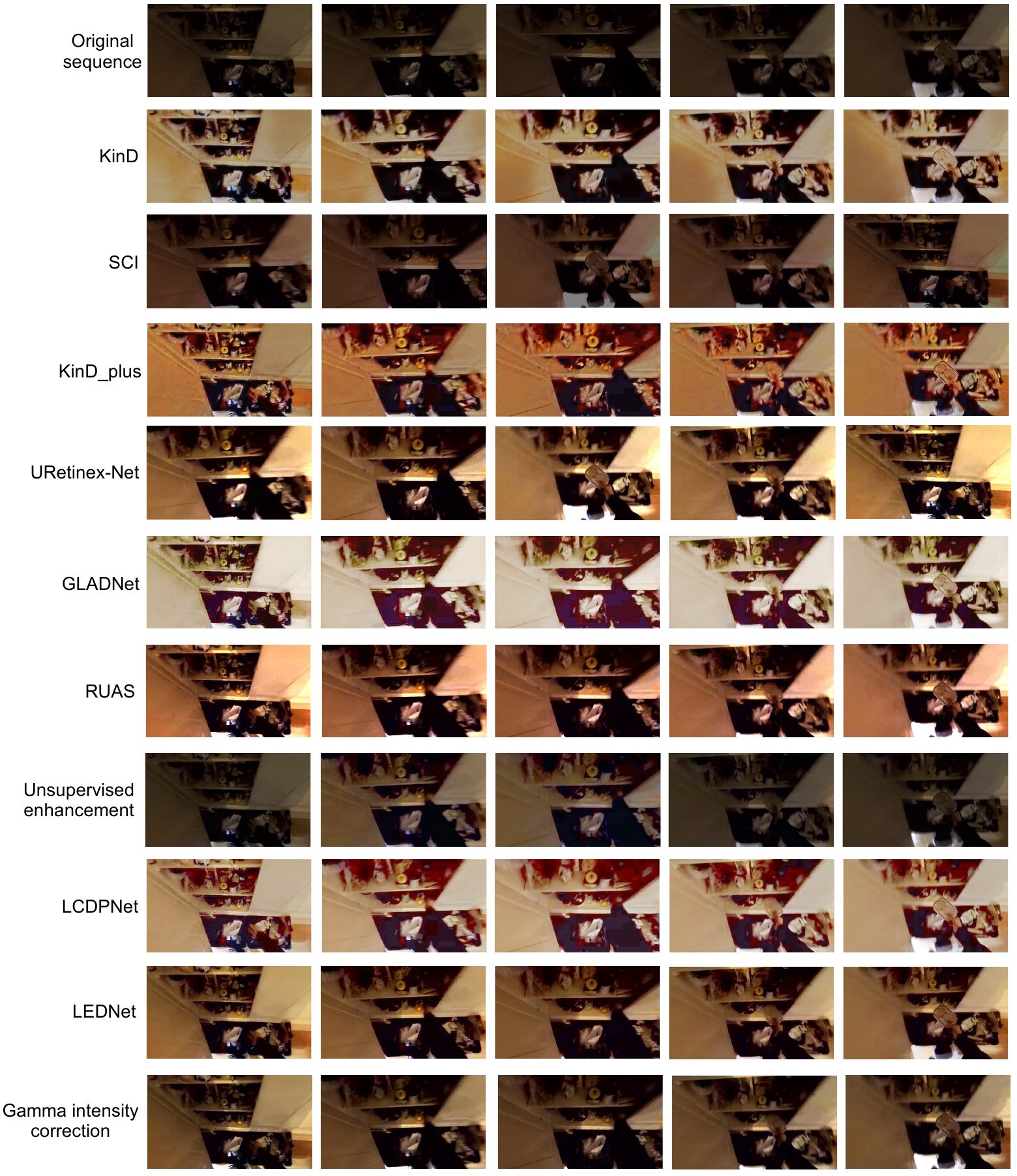}
\vspace{-0.5em}
\caption{\textbf{Video frames with image enhancement} for an example video with the \textit{take box} activity. Although image enhancers improve the illumination of dark frames, they result in color distortions that are harmful for activity recognition. 
}
\label{fig:image_enhancement}
\end{figure*}

\noindent \textbf{Visualization of Branch Attention. }
We visualize the branch attentions for several example video clips in Figure~\ref{fig:branch_att}. 
We observe that each type of branch attention tends to focus on different situations where certain features are more reliable. 
Specifically, the first branch has high response when the motion and sound are clearer than the appearance, while appearance, motion and sound are all useful when the second branch is activated. 
The activity appearance and sound are usually clear in videos with the activated third branch, and when only sound is reliable, the fourth branch has higher response. 
In addition, when few cues can be utilized, the fifth branch is often activated. 
Interestingly, the illuminance value $Y$ often cannot give a good indication of which features are reliable as videos of similar computed illuminance $Y$ can have various branch attentions. 
Thus, we conclude that the different branches allow our model to better utilize different features for recognition and adapt to various illumination conditions. 
We provide video examples in the supplementary video. 

\section{Visualization of Distortions}

We visualize the distortions introduced by the image enhancement methods in Fig.\ref{fig:image_enhancement}.
As we can see, the enhanced images are visually different from normal daylight video frames. Thus, the activity recognizer struggles to recognize the activities in these videos because of the distortions these methods add. 

\section{Failure Cases}

We provide failure cases in the supplementary video. 
Our model is prone to fail when there is a loud background sound similar to another activity class. In cases where different nouns can have a similar sound, our model may also predict the wrong noun when the visual features are unreliable in the dark. 
Our model also as similar failure cases to other activity recognizers. For instance, when a large proportion of video clip is not relevant to the activity class, on tail classes and when two activities are occurring at the same time.
We hope that these issues can be addressed in future works. 

\clearpage


\end{document}